%% file: main.tex
\documentclass[10pt,twocolumn,letterpaper]{article}

\usepackage{cvpr}              % To produce the CAMERA-READY version
\usepackage[accsupp]{axessibility}  % Improves PDF readability for those with disabilities.
\usepackage{url}
\usepackage{graphicx}
\usepackage{float}
\usepackage{array}
\usepackage{amssymb}
\usepackage{pifont}
\usepackage[table]{xcolor}
\usepackage{tabularx}
\usepackage{multirow}
\usepackage{booktabs}
\usepackage{adjustbox}
\usepackage{multirow}
\usepackage{booktabs}
\usepackage{makecell} % add this in preamble
\usepackage{float} % to fix figure location
\usepackage{tcolorbox}
\tcbuselibrary{skins, breakable}
\usepackage{xcolor}
\usepackage{listings}
\usepackage{amssymb}
\usepackage{pifont}
\usepackage{arydshln}
\usepackage{makecell}
\usepackage[normalem]{ulem}
\input{preamble}

\definecolor{cvprblue}{rgb}{0.21,0.49,0.74}
\usepackage[pagebackref,breaklinks,colorlinks,allcolors=cvprblue]{hyperref}

\def\paperID{15357} % *** Enter the Paper ID here
\def\confName{CVPR}
\def\confYear{2026}
\newcommand{\cmark}{\ding{51}} % ✓ plain black
\newcommand{\xmark}{\ding{55}} % ✗ plain black
\newcommand{\cxmark}{\ding{52}\rotatebox[origin=c]{-9.2}{\kern-0.7em\ding{55}}} % ✔✘
\newcommand{\tablescale}{0.8}      % <--- adjust once for all tables
\newcommand{\datasetname}{CRIT}
\title{CRIT: Graph-Based Automatic Data Synthesis to Enhance \\ Cross-Modal Multi-Hop Reasoning}

\author{
Junyoung Sung$^{1}$ \quad 
Seungwoo Lyu$^{1}$ \quad 
Minjun Kim$^{1}$ \quad
Sumin An$^{1}$ \quad \\
Arsha Nagrani$^{2}$ \quad 
Paul Hongsuck Seo$^{1}$ \\
$^{1}$Dept. of CSE, Korea University \quad
$^{2}$Google DeepMind \\
{\tt\small \{jys7451, dbtmddn41, ddomjun, suminan, phseo\}@korea.ac.kr}, \quad
{\tt\small arsha.nagrani@gmail.com}
}
\begin{document}
\maketitle
\input{sec/0_abstract}    
\input{sec/1_intro}
\input{sec/2_related_works}
\input{sec/3_methodology}
\input{sec/4_CRUX}

\input{sec/5_experiments}
\input{sec/6_error_analysis}

\input{sec/7_conclusion}
% \input{rebuttal}

{
    \small
    \bibliographystyle{ieeenat_fullname}
    \bibliography{main}
}
\clearpage
\input{sec/X_suppl}


\end{document}

%% file: preamble.tex
\newcommand{\supp}{Supp.~Mat.\xspace}

%% file: sec/0_abstract.tex
\begin{abstract}
Real-world reasoning often requires combining information across modalities, connecting textual context with visual cues in a multi-hop process. Yet, most multimodal benchmarks fail to capture this ability: they typically rely on single images or set of images, where answers can be inferred from a single modality alone. This limitation is mirrored in the training data, where interleaved image–text content rarely enforces complementary, multi-hop reasoning. As a result, Vision-Language Models (VLMs) frequently hallucinate and produce reasoning traces poorly grounded in visual evidence. To address this gap, we introduce {\datasetname}, a new dataset and benchmark built with a graph-based automatic pipeline for generating complex cross-modal reasoning tasks. {\datasetname} consists of diverse domains ranging from natural images, videos, and text-rich sources, and includes a manually verified test set for reliable evaluation. Experiments on this benchmark reveal that even state-of-the-art models struggle on such reasoning tasks. Models trained on {\datasetname} show significant gains in cross-modal multi-hop reasoning, including strong improvements on SPIQA and other standard multimodal benchmarks.
\end{abstract}

%% file: sec/1_intro.tex
\section{Introduction}

As humans, we are constantly interacting with a multimodal world. Real-world tasks often require cross-modal reasoning across multiple, interleaved sources of information. For instance, when following a DIY tutorial or a recipe illustrated with images, we may find ourselves continually cross-referencing textual instructions with a sequence of images, engaging in a multi-hop process to connect steps, tools, and outcomes ~\citep{webqa, unimmqa, m3docrag}. This involves grounding textual descriptions (e.g., “fold the dough”) to corresponding visual content (e.g., an image showing the dough being folded) and maintaining consistency of entities and states across modalities and steps.

Despite the importance of such tasks, most existing multimodal benchmarks do not adequately assess this capability. Current benchmarks, while valuable--typically present a single image ~\citep{chartqa, mmbench, mmvet, mmstar,mathvista}, a single video, or a static set of images ~\citep{muirbench, BLINK, MMIU}, with questions and answers represented in the textual domain. Even in datasets designed for multi-stage reasoning ~\citep{mathverse, neptune, minerva}, the necessary information can often be inferred from a single modality alone, failing to test true cross-modal grounding. In Figure~\ref{fig:teaser}, we provide an example of the complex reasoning we target. The model must execute multi-hop, cross-modal reasoning to reach the correct answer. It must first identify the conceptual topic `digital autonomy', which is associated with the student Elara Myles in the textual description, then determine from additional context that the `laptop' symbolizes digital autonomy, and finally locate the laptop within the image to conclude that its color is `silver'. Crucially, the multimodal information is \textit{complementary}; the visual data provides attribute information while the text provides additional context unavailable in the pixels alone.

This gap in evaluation is mirrored by a lack of suitable post-training or supervised fine-tuning (SFT) data. While a lot of interleaved image-text data is used during training, it is unclear how much of it is truly complementary, and hence how much the model is forced to correlate the two modalities. Given the scarcity of such complex interleaved data, it is perhaps  unsurprising that existing Vision-Language Models (VLMs) struggle on such complex reasoning tasks. As we show in our experiments, when prompted for step-by-step reasoning, Chain of Thought (CoT) traces output by models are often poorly grounded in the visual and textual evidence, frequently disjoint from the multimodal context and exhibiting significant hallucination.

\input{figures/task_figure}

In this work, we aim to address this problem by creating a scalable pipeline for generating high-quality, multi-hop, cross-modal reasoning data. Manually collecting such data is prohibitively expensive. While recent works have leveraged VLMs to scale up data collection~\citep{sharegpt4v, mammoth_vl, vision-r1, llava-cot, r1-onevision}, this approach has significant drawbacks for our target task. 
Tasking a VLM to automatically generate complex reasoning questions is prone to the same grounding failures and hallucinations we seek to measure. Furthermore, this risks creating cyclical biases, where models are evaluated on data generated by the very same class of models.

To overcome these challenges, we propose a novel graph-based automatic data generation pipeline for interleaved image-text content. Our pipeline is built on several key properties: (i) First, we use graphs as a structured representation of content, capturing entities, attributes, and relationships that appear in either modality. These are derived from reliable sources, such as manually annotated captions; (ii) Second, this structured format allows us to programmatically sample sub-graphs, guaranteeing the presence of complex, multi-hop relationships between modalities; and (iii) Finally, given a sampled sub-graph, we use a model to generate a complex question that necessitates multi-hop reasoning to be solved. By design, our pipeline does not require a VLM at any stage of the question-generation process (LLM is sufficient), thus avoiding the aforementioned cyclical biases and grounding issues.

Using this pipeline, we construct a novel dataset called \textbf{{\datasetname}} (\textbf{C}ross-modal multi-hop \textbf{R}easoning over interleaved \textbf{I}mage-\textbf{T}ext). Models trained on {\datasetname} show improved cross-modal multi-hop reasoning, while our test set--built by our automatic pipeline and refined by human annotators--serves as a reliable benchmark for this capability.

To summarize, our key contributions are as follows:
\begin{enumerate}
    \item We present an automatic graph-based data generation framework for cross-modal multi-hop reasoning across diverse domains, ranging from natural images to videos and text-rich sources such as scientific papers.
    \item We use this pipeline to create the {\datasetname} dataset where the test set has been verified manually by human raters. Along with the test set, we also benchmark a number of state-of-the-art VLMs and provide an in-depth analysis of common failure modes.
    % \item We further show that models trained on {\datasetname} achieve significant improvements in cross-modal multi-hop reasoning, with notable improvements on SPIQA and modest improvements on other multi-image benchmarks.
    \item We further show that models trained on {\datasetname} achieve significant gains in cross-modal multi-hop reasoning, with notable improvements on SPIQA and modest improvements on other standard multimodal benchmarks.
\end{enumerate}

%% file: figures/task_figure.tex
\begin{figure*}[t]
\centering
\includegraphics[width=1\textwidth]{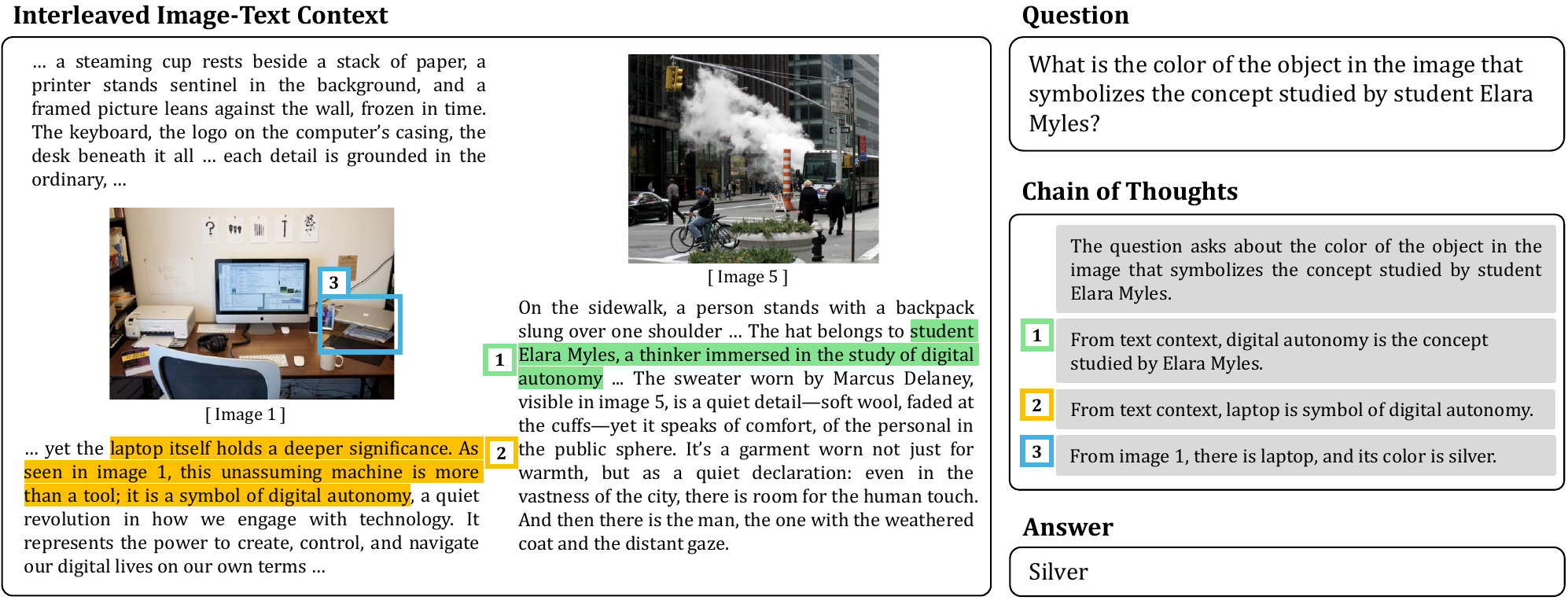}
\caption{\textbf{Example from CRIT.} 
The CRIT dataset is designed to train and evaluate cross-modal multi-hop reasoning over interleaved image-text contexts. Each example combines multiple textual and visual sources that must be jointly interpreted to answer a compositional question. The colored highlights in the text indicate distinct evidence segments contributing to the reasoning chain, while the numbered boxes in the images and text correspond to the multi-step inference process outlined in the Chain of Thoughts panel. Together, they illustrate how CRIT requires connecting dispersed clues across modalities to reach a grounded answer.}
\label{fig:teaser}
\end{figure*}

%% file: sec/2_related_works.tex
\section{Related Work}
%\input{iclr2026/tables/comparison_with_existing_work}
%\subsection{Vision Language Models}
\noindent\textbf{Cross-Modal Multi-Hop Reasoning} \ \ 
Benchmarking cross-modal multi-hop reasoning has received considerable attention in recent years \citep{webqa, mumuqa, ct2c, fm2ds, mmcv, tanq, wikimixqa, ii-mmr}. While these benchmarks aim to assess a model’s ability to integrate information across modalities through multiple reasoning steps, many fall short of capturing the full complexity of the task. Datasets such as MMQA~\citep{multimodalqa}, for instance, are often solvable without visual inputs, indicating weak multimodal dependencies \citep{fcmr}. 
FCMR~\citep{fcmr} introduces more challenging settings but still provides only QA pairs without explicit reasoning traces and remains limited in scope, focusing solely on the financial domain. Other datasets rely heavily on Wikipedia content, lack training splits, or depend on manual construction which limits scalability. Importantly, prior work seldom investigates cross-modal multi-hop reasoning over interleaved image–text inputs, where both modalities must be identified and combined to locate relevant evidence and answer questions accurately. To address these limitations, we introduce CRIT, with a detailed comparison to existing cross-modal multi-hop datasets provided in \supp

\noindent\textbf{Interleaved Image-Text Datasets} \ \ 
%Interleaved image–text datasets \citep{mmc4, obelics, omnicorpus, mm_textbook} are now standard for pretraining VLMs, but their weak image–text alignment limits their usefulness for building datasets that require cross-modal multi-hop reasoning. Although such pretraining improves general multimodal ability, it does not ensure strong performance on tasks requiring integration of complementary information across modalities.
%Recently, datasets such as SPIQA~\citep{pramanick2024spiqa} and VEGA~\citep{vega} have been proposed to target interleaved image–text comprehension. SPIQA reports performance under single-modality settings and achieves relatively strong results, indicating that models can perform well without combining modalities — suggesting limited cross-modal dependency. VEGA constructs interleaved contexts and QA pairs using SciGraphQA \citep{scigraphqa}, but much of the information used to create SciGraphQA’s QA pairs remains in VEGA's textual context, making many questions answerable from text alone.
%While performance on these datasets is still meaningful, real-world multimodal settings frequently involve complementary, non-redundant information spread across multiple images and passages. Capturing such richer interactions requires datasets that move beyond weakly aligned interleaved corpora.
Interleaved image–text datasets \citep{mmc4, obelics, omnicorpus, mm_textbook} are widely used for pretraining VLMs, but their weak cross-modal alignment limits their effectiveness for tasks requiring multi-hop reasoning across modalities. While they improve general multimodal capabilities, they do not guarantee strong performance when integrating complementary information from both modalities.
Benchmarks such as SPIQA~\citep{pramanick2024spiqa} and VEGA~\citep{vega} attempt to address interleaved comprehension. However, SPIQA shows that models can perform well using only a single modality, suggesting limited cross-modal dependency. Similarly, VEGA builds on SciGraphQA \citep{scigraphqa}, but much of the necessary information remains in the textual context, allowing many questions to be answered without visual input.
Although these datasets remain useful, real-world scenarios often involve non-redundant, complementary information distributed across modalities. This highlights the need for datasets that better capture strong cross-modal interactions beyond loosely aligned interleaved corpora.

%% file: sec/3_methodology.tex
\section{Data Generation for Cross-Modal Multi-Hop Reasoning}
% \section{{\datasetname}: A Cross-Modal Reasoning Data Generation Framework}
\label{sec:method}
\input{figures/3_main_figure}
% \subsection{Overview}
In this work we explore cross-modal multi-hop reasoning as shown in Figure~\ref{fig:teaser}.
% In particular, we investigate whether VLMs can answer questions that require connecting multiple facts distributed across both modalities in image–text interleaved content, and reasoning over these connected facts.
Obtaining suitable data for this task is non-trivial--even prior to annotation, the first challenge is identifying source data that contains high-quality image–text interleaved content with rich multimodal interactions. Once such content is collected, annotation requires identifying connected facts across modalities, formulating multi-hop questions and answers based on these facts, and ensuring that the supporting facts lead to a unique answer. One potential way to create such annotations is to prompt LLMs to automatically generate multi-hop questions and answers from raw image–text content. In practice, however, directly prompting the LLM to produce such examples without additional structure rarely yields valid cross-modal multi-hop data. The model often generates questions that rely on only one modality, collapse multi-hop reasoning into a shallow single-hop relation, or introduce hallucinated facts that are not grounded in the source content. We provide an illustrative failure case in \supp

%One technique that has recently been adopted to reduce this annotation burden is leveraging pretrained LLMs or VLMs with prompting for annotation or data synthesis. However, such approaches often suffer from limited reliability, lack of controllability, and inherent hallucination.
%To address these challenges, we propose a novel graph-based framework for automatic multimodal QA data generation tailored to cross-modal multi-hop reasoning. 
%Our approach leverages existing annotated images with scene graphs and generates coherent textual contexts, thereby constructing interleaved image–text content suitable for reasoning tasks.
%In this formulation, visual and textual content are represented as a graph where entities, attributes, and relationships are explicitly modeled. 
%This structured representation provides precise control over information from each modality, allows us to enforce cross-modal reasoning, and exposes explicit reasoning traces that can be leveraged for QA and chain-of-thought generation.
%Our framework consists of three main stages, as illustrated in Figure~\ref{fig:main}:
%(a) multimodal content graph construction, 
%(b) textual context generation, and 
%(c) question–answer generation, which we describe in detail below.

To address these challenges, we introduce a novel graph-based automatic data generation framework.
Our framework operates in three stages, illustrated in Figure~\ref{fig:main}:
\textbf{(1) constructing multimodal content graphs},
\textbf{(2) generating textual context} to accompany visual content, and
\textbf{(3) generating question–answer pairs} that require cross-modal multi-hop reasoning based on the multimodal content graph.
In the following subsections, we describe each stage in detail. We first describe how our framework works with images that already contain annotated scene graphs, and then describe extensions to our pipeline that enable us to use videos and scientific papers as well (Sec. \ref{sec:extension}).

\subsection{Multimodal Content Graph Construction}
\label{sec:graph_construction}

The multimodal content graph lies at the core of our framework. It provides a structured representation of interleaved image–text content, consisting of entities, attributes, and relationships that appear in either modality. The graph encodes both intra-modal and cross-modal relationships.

\noindent\textbf{Formulation of Multimodal Content Graphs.} \ \ 
We model visual and textual content in interleaved image–text as a directed graph $G = (\mathcal{V}, \mathcal{E})$. 
Each node $v \in \mathcal{V}$ represents an entity, either a visual object in an image or a textual entity. 
An edge $e \in \mathcal{E}$ between nodes $u$ and $v$ is represented as a triple $(u, v, r)$, where $r$ denotes the relation between the two entities.
The graph is further augmented with attributes associated with each entity when available.
%Finally, a modality function $m(\cdot)$ specifies the modality index: for an entity or relation, $m(\cdot) = i$ indicates presence in image $i$, while $m(\cdot) = 0$ denotes presence in the text.
%\phseo{never used?1}
% Each node in \(V\) is a triple \((e, s, a)\), where \(e\) is an entity name, \(s \in \{\texttt{text}, \texttt{image}\}\) denotes the source, and \(a\) is a set of attributes serving as visual descriptions when available. Nodes are uniquely determined by \((e, s)\). If multiple instances of the same entity appear in an image, we simply treat them as distinct nodes by assigning instance-specific identifiers. Each edge in \(E\) is written as \((u, v, r, \ell)\), where \(u, v \in V\) are distinct nodes, \(r\) is the semantic relation type, and \(\ell\) marks the modality source. Relations extracted from image \(i\) carry the label \(\ell = \texttt{image}_i\), whereas those coming from text or cross-modal alignment use \(\ell = \texttt{text}\).

\noindent\textbf{Graph Construction through Content Augmentation.} \ \ 
We begin with a dataset of images annotated with scene graphs, eliminating the need for image generation or potential errors in scene-graph recognition.
Since our target is interleaved image–text content involving multiple images, we randomly sample one to six images from the annotated dataset (Figure~\ref{fig:main} (a)).
For each image, we use a rule-based method to retain only entities that can be uniquely identified by an attribute or by their relation to another entity, ensuring that each attribute or relation corresponds to exactly one entity (Figure~\ref{fig:main} (b)).
This avoids ambiguity during the QA generation phase, where multiple valid answers could otherwise arise.
The scene graphs of the sampled images are then merged into a single content graph, which serves as the starting point for our framework.
To augment this graph with textual information, we leverage the LLM to add new entities and relations.
Specifically, for each image node in the graph, we prompt the LLM to generate a plausible entity and relation connected to that node (Figure~\ref{fig:main} (c)).
We then apply an additional round of prompting to create plausible relations among the newly added entities (Figure~\ref{fig:main} (d)). 
These augmented entities and relations constitute the textual content; since they are not tied to a specific image, they serve as bridges that connect entities across different images through related textual nodes.
The resulting graph provides a coherent multimodal context for evaluating cross-modal multi-hop reasoning, as it connects concepts drawn from multiple images and textual segments.

\subsection{Textual Context Generation}
\label{sec:textual_context_generation}

After graph construction, each sample is represented by a unified graph structure, comprising both image and text nodes. For each image, we prompt the LLM with a subgraph to generate complementary textual context. 
We begin by extracting the subgraph that includes the corresponding image nodes and all text nodes directly connected to them, along with the edges between those text nodes. When an edge connects two text nodes associated with different images, we randomly assign the edge to one image and expand its subgraph accordingly (Figure~\ref{fig:main} (e)). 
%Including these edges is essential because they provide the only connections across independent images, enabling the construction of multi-hop reasoning questions and supporting cross-modal reasoning that spans multiple images (Figure~\ref{fig:main}e). 
Attributes of image nodes and inter-image relations are excluded, as these are intended to be inferred from the image during cross-modal reasoning. The generated passages therefore describe only the augmented text nodes, the relationships between them and their connections to the image nodes (Figure~\ref{fig:main} (f)). Diverse narrative styles—such as stories, diary entries, or documentary-style scripts are used to provide linguistic variety. By combining the generated texts with the images, the framework yields complementary multimodal content that facilitates cross-modal multi-hop reasoning.

\subsection{Question-Answer Pair Generation}
\label{sec:qa_generation}
Once the multimodal content is generated by augmenting images with textual content, we proceed to generate cross-modal multi-hop QA pairs. 
To this end, we sample a subgraph from the multimodal content graph that represents a chain of facts within the content (Figure~\ref{fig:main} (g)). 
Since our goal is cross-modal reasoning, we ensure that the sampled subgraph includes nodes from both images and text. 
To guarantee multi-hop reasoning, we further restrict the subgraphs to have $1 \leq h \leq 5$ edges, such that answering the resulting questions requires reasoning over multiple connected facts.
Finally, we require the terminal node in each chain to come from an image. 
For $h=1$, we sample such subgraphs only when the terminal image node possesses an attribute and use this attribute as the answer to ensure the reasoning remains two-hop.
For $h>1$, we select either its entity name or one of its attributes as the answer, grounding the final reasoning step in the image content.
This strategy improves the reliability of evaluation by preventing the model from guessing answers based solely on textual biases without performing the correct cross-modal reasoning. 

Given a sampled subgraph (serialized as JSON) and its designated answer, we prompt the LLM to generate a question under the following constraints:
(i) the answer to the generated question must match the provided target answer;
(ii) solving the question must require reasoning over the entire chain of facts in the subgraph, thereby enforcing cross-modal multi-hop reasoning; and
(iii) intermediate entities should not be mentioned explicitly in the question but instead be recoverable only through multi-hop inference (Figure~\ref{fig:main} (h)). 
While the QA pair provides the basis for evaluating cross-modal multi-hop reasoning, we additionally generate a CoT trace~\citep{wei2022chain} to be used for training and evaluation.
Since the subgraph corresponding to each QA already encodes the complete reasoning chain, we can leverage it to prompt the LLM to produce a CoT response for the generated question.
This offers extra supervision by explicitly indicating where each piece of information should be retrieved from within the multimodal context.

Once the QA samples with CoT are generated, we apply the LLM-based staged filtering process to ensure quality. 
First, we discard questions that explicitly mention intermediate entities from the sampled subgraph, thereby ensuring multi-hop reasoning by eliminating potential shortcuts in the reasoning chain. 
Next, we filter out questions that can be answered using a single modality.
To perform this test, we provide LLMs with the nodes and edges corresponding to each modality; 
since visual content is translated into structured text through the graph representation, a text-only LLM is sufficient. 
We use three different LLMs and remove a question if all three predict the correct answer using only one modality (either text or visual).
Finally, we prune CoT responses that are excessively verbose by limiting the output to a maximum of ten sentences.

\subsection{Extensions to Videos and Text-Rich Sources}
\label{sec:extension}
The data creation pipeline described above is limited in two aspects: (i) first, it relies on images that have graph annotations, and (ii) the images are independent and unrelated to one another, as well as being limited to the natural image domain. We therefore aim to extend the pipeline to settings lacking pre-existing graph information by first deriving graph representations from alternative annotations and then applying the same pipeline. We also seek to incorporate text-rich visuals such as tables, diagrams, and charts that frequently occur alongside textual context in real-world documents ~\citep{leopard, mplug-docowl-1.5, llavar}. 
To address these gaps, we extend our pipeline to two additional data sources: 
(i) still frames from videos, where multiple coherent frames naturally share contextual information, 
and (ii) scientific papers, which provide abundant text-rich images.

For video frames, the main challenge is the absence of scene-graph annotations. 
To address this, we leverage dense video captioning datasets, where long-form videos are annotated with temporally localized event descriptions. 
We select the frames with the highest CLIP similarity ~\citep{clip} to each caption to ensure that the visual content aligns closely with the description. 
Although these captions may be short, they provide sufficient content information to be converted into partial scene graphs—similar to the manual scene graph annotation process in ~\citep{visual_genome}, where captions are collected and then transformed into graphs.
Concretely, we prompt the LLM to convert these descriptions into graphs.
Sampled frames from a single video may share entities and relations, and therefore we incorporate this into both the prompting design and the resulting graphs for coreference resolution.
After constructing a scene graph across coherent frames, the rest of the pipeline proceeds as before, except that the textual context is now generated once for the entire image set rather than separately for each image.

In the case of text-rich images, scientific papers inherently provide high-quality interleaved image–text content. 
Because well-aligned textual descriptions already exist, generating complementary textual context is unnecessary. 
Nonetheless, constructing QA pairs would benefit from a graph-based representation. 
To this end, we convert the multimodal content—paragraphs, figures, and tables—into a unified graph structure, treating both figures and rendered tables as images.
Given a paper containing textual content with associated figures and tables, we first construct a content graph from paragraphs that do not reference any visual elements using the LLM. For paragraphs that reference figures or tables, we prompt the LLM to identify visual entities and relations. Numerical entities, comparisons, and explicit visual descriptions are elevated to visual nodes in the graph.
To enforce cross-modal reasoning, the corresponding sentences are then removed from the text body once they are marked as visual.
Finally, once the corresponding multimodal content graph is constructed, QA generation is performed following the same base pipeline. In this case, we restrict the subgraph to have $1 \leq h \leq 4$ edges, as their relations are more complex and larger chains often lead the LLM to generate overly complicated or unnatural questions. Further implementation details and prompt templates can be found in \supp

%% file: figures/3_main_figure.tex
\begin{figure*}[t]
\centering
\includegraphics[width=1\textwidth]{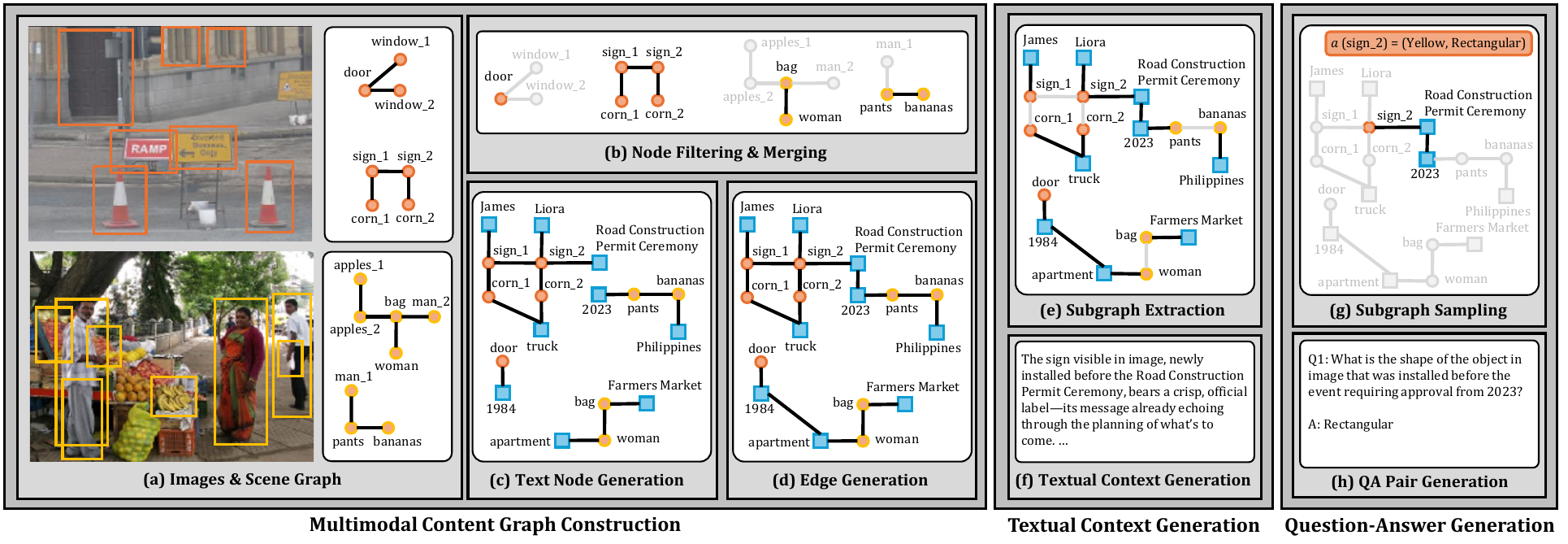}
\caption{\textbf{Overall Process of Cross-Modal Multi-Hop QA Generation.} 
The procedure consists of three main stages. 
\textbf{(1) Multimodal Content Graph Construction:} images annotated with scene graphs are sampled (a), unique entities are filtered and merged (b), and text nodes connected with edges are generated via LLM prompting (c, d). 
\textbf{(2) Textual Context Generation:} subgraphs are extracted from the multimodal content graph (e) to produce complementary textual descriptions (f). 
\textbf{(3) Question–Answer Generation:} subgraphs are further sampled (g) to generate QA pairs requiring cross-modal multi-hop reasoning (h). 
Orange and yellow circular nodes represent visual entities originating from different images, while blue square nodes denote text nodes. 
Entities with identical names are distinguished by numerical subscripts (e.g., \textit{apple\_1}, \textit{apple\_2}).}

\label{fig:main}
\end{figure*}

%% file: sec/4_CRUX.tex
\section{{\datasetname}}
In this section, we describe the implementation details used to create {\datasetname} (sec. \ref{sec:implementation_details}), and dataset statistics (sec. \ref{sec:dataset_statistics}). 

\subsection{Data Sources and Models} \label{sec:implementation_details}

\noindent\textbf{Data Sources} \ \ 
{\datasetname} is based on three primary data sources, corresponding to three distinct domains: natural image, video frame, and scientific paper. Specifically, the natural image domain is constructed from GQA~\cite{hudson2019gqa}, the video frame domain from ActivityNet Captions~\cite{krishna2017dense}, and the scientific paper domain from SPIQA~\cite{pramanick2024spiqa}. We follow the official train/test splits of each source to create the {\datasetname} training set and benchmark. From GQA, we utilize images and their associated scene graphs, sampled and processed following the pipeline described in Section~\ref{sec:graph_construction} to Section~\ref{sec:qa_generation}. Beyond these, {\datasetname} additionally incorporates video frames and captions from ActivityNet Captions, and also includes interleaved image–text pairs extracted from scientific papers in SPIQA, as described in Section~\ref{sec:extension}. Importantly, no QA pairs from these datasets are directly used in {\datasetname}; all QA annotations are newly generated through our pipeline. Further implementation details and all prompts used are included in \supp

\noindent\textbf{Models} \  \
Qwen3-30B-A3B-Instruct-2507~\citep{qwen3} was used at every stage of the data generation pipeline. 
For the final filtering of questions and answers, we used 
Qwen3-30B-A3B-Instruct-2507~\citep{qwen3}, 
Gemma-3-27b-it~\citep{gemma3}, 
and Mistral-Small-3.2-24B-Instruct~\citep{mistral2025small3_1}.

\input{tables/data_statistics}

\subsection{Dataset Analysis} \label{sec:dataset_statistics}
{\datasetname} is divided into train and test sets derived from the official splits of each data source. For the SPIQA source, due to the limited size of its official test set, we sample a subset from its training split to form the {\datasetname} test split for the scientific paper domain, ensuring no overlap between train and test samples. For the test split, we recruited 12 English-proficient annotators majoring in STEM disciplines to manually verify and filter samples that did not adhere to the guidelines. In contrast, the train set is automatically constructed without human curation, and may therefore contain some degree of noise.
Each {\datasetname} data sample comprises a set of one or more images, an accompanying textual context, one or multiple QA pairs, and corresponding CoT reasoning traces. During training, samples with multiple QA pairs are concatenated to form multi-turn conversations while test samples are restricted to single-turn QA pairs.
In total, {\datasetname} contains 84,199 train samples and 1,446 test samples, corresponding to 269,467 QA pairs in the train split and 1607 QA pairs in the test split. Table~\ref{tab:data_stat} summarizes the detailed statistics across data sources.

%Each sample consists of multiple QA pairs per image set, concatenated into a multi-turn conversation. In total, {\datasetname} consists of 84,179 samples in training and 1,988 in testing, and 249,442 QA pairs in the training set and 2,207 QA pairs in the test set.
%Table~\ref{tab:data_stat} presents the statistics for each data source. Note how the majority of QA pairs are generated from natural images annotated with scene graphs. Some QA pairs also include image-image reasoning in addition to image–text reasoning. This proportion varies across the different domains. For natural images consisting of independent images, this proportion is 0\% (as none of the images are related).
%For videos composed of sequential frames, the proportion is highest, at 21.6\% in the training set and 34.0\% in the test set.
%For scientific papers, the corresponding proportions are 6.1\% and 5.3\%.

%% file: tables/data_statistics.tex
\begin{table}[t]
\centering
\caption{\textbf{{\datasetname} Dataset Statistics.} 
{\datasetname} comprises three domains: natural image (NI), video frame (VF), and scientific paper (SP). 
This table reports the key statistics for each domain.}
\scalebox{\tablescale}{
\begin{tabular}{llccc}
\toprule
\textbf{Split} & \textbf{Statistic} & \textbf{NI} & \textbf{VF} & \textbf{SP} \\
\midrule

\multirow{8}{*}{Train} 
& \# Samples & 49,159 & 8,022 & 27,018\\
& Avg. Images/Sample & 3.8 & 3.5 & 3.7 \\
& Avg. Text Tokens/Sample & 2,442 & 649 & 5,443 \\
& \# QA & 153,781 & 16,071 & 99,615 \\
& \quad \# 2-hop & 109,735 & 8,061 & 58,429 \\
& \quad \# 3-hop & 12,271 & 6,042 & 28,385 \\
& \quad \# 4-hop & 12,592 & 849 & 12,801 \\
& \quad \# 5-hop & 19,183 & 1,119 & -- \\

\midrule

\multirow{8}{*}{Test} 
& \# Samples & 792 & 332 & 322\\
& Avg. Images/Sample & 4.4 & 5.2 & 7.7 \\
& Avg. Text Tokens/Sample & 2,930 & 802 & 5,338 \\
& \# QA & 858 & 366 & 383 \\
& \quad \# 2-hop & 597 & 134 & 245 \\
& \quad \# 3-hop & 71 & 163 & 110 \\
& \quad \# 4-hop & 74 & 34 & 28 \\
& \quad \# 5-hop & 116 & 35 & -- \\

\bottomrule
\end{tabular}
}
\label{tab:data_stat}
\end{table}

%% file: sec/5_experiments.tex
\section{Experiments}

\input{tables/results_on_crux_refined_cot}
\input{tables/cross_modal_benchmark_results}
\input{tables/standard_multimodal_benchmark_result}

\subsection{Experimental Setup}
\noindent\textbf{Base Models}  
We evaluate both open-source and proprietary models. For open-source models, we evaluate Qwen2.5-VL~\citep{qwen2-5-vl}, LLaVA-Onevision~\citep{llava_onevision}, Idefics2~\citep{idefics2}, Intern2.5-VL~\citep{internvl-2.5}, Phi3.5-Vision~\citep{phi}. For proprietary models we use GPT-4o, GPT-4o-mini~\citep{gpt-4o} and Gemini-2.0-flash \citep{gemini}.

\noindent\textbf{Training Settings}  
We perform supervised fine-tuning using LoRA~\citep{lora} on VLMs that support multi-image input, specifically Qwen2.5-VL-7B and Idefics2-8B on the {\datasetname} training set.
% Each fine-tuned models are notated as Qwen2.5-VL$_{\text{{\datasetname}}}$ and Idefics2$_{\text{{\datasetname}}}$.
Each training sample is included in two formats: one containing only the final answer and another including the full CoT reasoning trace. We use both formats to enable the models to generate concise answers as well as step-by-step reasoning responses.

\noindent\textbf{Evaluation Details}  
Evaluation is conducted on the {\datasetname} benchmark, which covers three domains: natural images, videos, and scientific papers. Two evaluation settings are considered: (1)~\textit{Direct Answer}, where the model is prompted to produce an answer directly without reasoning steps, and (2)~\textit{CoT}, where the model is instructed to reason step-by-step before providing the final answer. Model performance is measured using Exact Match (EM) and F1 scores on the final answer~\citep{squad}. We observe similar trends across both evaluation settings; therefore, unless otherwise specified, results are reported using CoT prompting by default, with direct answer results provided in \supp

%Evaluation spans natural images, video, and scientific papers \phseo{?}, with Exact Match (EM) and F1 as metrics. \phseo{for cot results?}We also report \textit{Reference Accuracy}, measuring whether the model correctly identified the necessary images to answer the question. All prompts use CoT\phseo{?}, and models are instructed to explicitly reference the relevant image\phseo{?}.

\subsection{Results on {\datasetname}}
Table~\ref{tab:{\datasetname}_refined_cot_result} presents the performance of various VLMs on the {\datasetname} benchmark under the CoT evaluation setting.
Overall, all models perform poorly, including proprietary systems. Although proprietary or larger open-source models such as GPT-4o and Qwen2.5-VL-72B achieve moderately better results than smaller models, their absolute performance remains low.
This reveals that, despite recent advances in multimodal reasoning, existing models still have limited capability in cross-modal multi-hop reasoning, highlighting the substantial challenges posed by {\datasetname}.
When fine-tuned on the {\datasetname} training set, both Qwen2.5-VL$_\text{{\datasetname}}$ and Idefics2$_\text{{\datasetname}}$ demonstrate substantial improvements over their corresponding baselines across all domains, surpassing even proprietary models in both EM and F1.
Nonetheless, the scientific-paper domains remain particularly challenging.
For scientific papers, long and complex textual contexts hinder precise alignment between visual and textual information. As a result, even trained models continue to struggle with maintaining coherent reasoning in these settings. Model performance across different hop counts is included in Supp. Mat.

%%Table~\ref{tab:{\datasetname}_result} reports results across natural images, videos, and scientific papers. Open-source models without fine-tuning perform poorly. They often fail to localize the relevant image or, even when successful, provide incorrect answers due to visual perception errors or the inability to connect information across modalities.
%Fine-tuning on {\datasetname} leads to substantial gains across all domains. Qwen2.5-VL\(_{\text{{\datasetname}}}\) achieves the best EM and F1 score for natural images, its reference accuracy reaching the level of proprietary models. Idefics2$_{\text{{\datasetname}}}$ which didn't have the capability to find the relevant sources to answer the question improved substantially. The relatively high scores in the video domain for LLaVA-OneVision and Idefics2 were obtained while they were instructed to engage in reasoning, they provided short answers that fortunately turned out correct.

% metric 명시하거나 아니면 supple에 있다고 쓰기
\subsection{Results on Other Benchmarks}
\label{sec:results_on_other_benchmark}

We further investigate the impact of fine-tuning on {\datasetname} across several benchmarks that evaluate the general multimodal capabilities of VLMs, including SPIQA~\citep{pramanick2024spiqa}, VEGA~\citep{vega}, MMQA~\citep{multimodalqa}, FCMR~\citep{fcmr}, MuirBench~\citep{muirbench}, BLINK~\citep{BLINK}, MP-DocVQA~\citep{mp-docvqa}, MMStar~\citep{mmstar}, MME~\citep{mme}, SeedBench~\citep{seedbench} and ChartQA~\citep{chartqa}.
The first four benchmarks primarily focus on complex cross-modal reasoning, while the remaining benchmarks represent standard multimodal evaluation suites covering general capabilities across both single-image and multi-image settings.
All evaluations follow the official settings and metrics defined for each benchmark.
Specifically, we compare the pretrained Idefics2 model and its variants fine-tuned with and without {\datasetname}, in addition to Mantis-Instruct~\citep{mantis}, across these benchmarks.
Mantis-Instruct is a widely adopted public dataset known for improving general instruction-following capabilities over multiple images in VLMs. 
We jointly fine-tune on Mantis-Instruct and CRIT to preserve the model’s existing instruction-following and multimodal capabilities while allowing CRIT to act as a complementary dataset that provides focused supervision for cross-modal reasoning.

Results on cross-modal reasoning benchmarks are reported in Table~\ref{tab:cross_modal_benchmarks}. These benchmarks are derived from diverse human-authored sources, such as academic publications, Wikipedia articles, and financial reports, reflecting a wide range of real-world reasoning scenarios. The pretrained model without fine-tuning performs poorly, reflecting limited exposure to cross-modal reasoning during pretraining. Fine-tuning on Mantis-Instruct alone improves performance, consistent with prior work~\cite{mantis}, while incorporating {\datasetname} yields substantial gains across all benchmarks. On SPIQA, fine-tuning with {\datasetname} more than doubles performance across METEOR, ROUGE-L, and CIDEr. Beyond SPIQA, training with CRIT yields consistent improvements across other benchmarks, including relative gains of 19\%/39\% (ROUGE-L/BLEU) on VEGA, 10\% (EM/F1) on MMQA, and 12\% (F1) on FCMR, demonstrating substantial improvements in general cross-modal reasoning. Note that although scientific paper data from SPIQA are used as contextual examples during dataset construction, no SPIQA question–answer pairs are included in either data construction or fine-tuning, and the SPIQA validation and test sets are never used at any stage.

Table~\ref{tab:mm_benchmark_comparison} presents results on broader multimodal benchmarks. Fine-tuning with {\datasetname} preserves performance on MME and SeedBench, while improving over the baseline on BLINK, MP-DocVQA, MMStar, and ChartQA. Performance on MuirBench remains comparable to the model trained with Mantis-Instruct alone.
Overall, {\datasetname} consistently enhances performance across both specialized cross-modal reasoning benchmarks and general multimodal evaluations. These findings suggest that training on more challenging cross-modal multi-hop reasoning examples improves the generalization and transferability of VLMs without degrading their existing capabilities.

\input{tables/results_data_scaling}

%\input{figures/task_figure}

% Further data creation 이런 느낌의 제목으로 바꾸기
\subsection{Extending the {\datasetname} Dataset}
We study whether scaling {\datasetname} improves performance by augmenting its training data, based on fine-tuning Qwen2.5-VL-7B. 
% The original dataset relies on human-annotated sources (e.g., GQA scene graphs and ActivityNet Captions), which are high-quality but limited in scalability. To address this, we construct additional samples using off-the-shelf models, introducing noisier but scalable annotations, and evaluate their impact on cross-modal reasoning.
The original {\datasetname} dataset relies on human-labeled data sources in the natural image and video frame domains, where scene graphs from GQA and captions from ActivityNet Captions are manually annotated sources. As a more scalable alternative, we extend {\datasetname} by constructing additional training samples using annotations produced by off-the-shelf models, which are inherently noisier than human-curated ones. This allows us to examine whether such noisy data can complement manually annotated samples and enhance cross-modal reasoning performance. 

To this end, we augment the {\datasetname} training set in both natural image and video domains. For natural images, we use images from Open Images~\citep{open_images} and generate scene graphs using EGTR~\cite{im2024egtr}, supplemented by object attributes generated using Qwen2.5-VL-7B. For videos, we use video frames extracted from Charades~\citep{charades}, DiDeMo~\citep{didemo}, and VidOR~\citep{vidor}, and generate video captions using Qwen2.5-VL-32B. Importantly, the same {\datasetname} data generation pipeline is applied to these automatically annotated sources, ensuring consistency with the original data generation pipeline. This augmentation produces approximately 100K additional samples for natural images and 25K for videos. 

% As shown in Table~\ref{tab:data_scaling}, augmentation consistently improves performance across domains, with the largest gains in the video frame domain, likely due to its smaller original dataset, which benefits more from additional samples. Performance also improves in the scientific paper domain despite no direct augmentation, indicating cross-domain transfer. These results demonstrate that model-generated annotations provide a scalable and effective way to enhance cross-modal reasoning.

As shown in Table~\ref{tab:data_scaling}, augmentation consistently improves performance across all domains. The largest improvement appears in the video domain, likely due to its smaller original dataset, which benefits more from additional samples. The performance of the scientific paper domain also improves even though augmentation is applied only to the natural image and video domains, suggesting cross-domain transfer of cross-modal reasoning capabilities. Overall, these results demonstrate that model-generated annotations provide a scalable and effective way to enhance cross-modal reasoning.

%% file: tables/results_on_crux_refined_cot.tex
\begin{table}[t]
\centering
\caption{\textbf{Results on {\datasetname}.}
Performance comparison of proprietary, open-source, and fine-tuned VLMs across natural image (NI), video frame (VF), and scientific paper (SP).
Models fine-tuned on {\datasetname} are highlighted with a gray background.
Open-source models are grouped by model size for clearer comparison.
\textit{\# Prm} denotes the number of model parameters.}
\setlength{\tabcolsep}{5pt} % Reduce horizontal padding between columns
\scalebox{\tablescale}{{\begin{tabular}{lccc ccc ccc}
\toprule
 & 
& \multicolumn{2}{c}{\textbf{NI}} 
& \multicolumn{2}{c}{\textbf{VF}} 
& \multicolumn{2}{c}{\textbf{SP}} \\
\cmidrule(lr){3-4} \cmidrule(lr){5-6} \cmidrule(lr){7-8}
\textbf{Model}&\textbf{\# Prm} & EM & F1 & EM & F1 & EM & F1 \\
\midrule

\multicolumn{8}{c}{\textbf{\textit{Proprietary Models}}} \\
\midrule
Gemini 2.0 Flash \citep{gemini}       & - & 30.4 & 32.7 & 30.6 & 35.3 & 11.5 & 13.7 \\
GPT-4o \citep{gpt-4o}       & - & 35.1 & 37.7 & 32.0 & 38.9 & 8.4 & 14.0 \\
GPT-4o-mini \citep{gpt-4o}   & - & 24.7 & 25.5 & 28.7 & 33.3 & 9.7 & 13.1 \\
\midrule

\multicolumn{8}{c}{\textbf{\textit{Open-Source Models}}} \\
\midrule
Phi3.5-Vision \citep{phi}    & 4B  & 21.5 & 23.0 & 24.0 & 26.5 & 3.1 & 5.9 \\
\cdashline{1-8}
LLaVA-Onevision \citep{llava_onevision} & 7B  & 31.0 & 32.0 & 33.3 & 37.4 & 7.1 & 10.1 \\
Qwen2.5-VL \citep{qwen2-5-vl}       & 7B  & 28.3 & 29.1 & 24.0 & 27.8 & 6.8 & 9.6 \\
\rowcolor{gray!30} \textbf{Qwen2.5-VL\(_{\text{{\datasetname}}}\)} & 7B & \textbf{58.6} & \textbf{59.5} & \textbf{38.8} & \textbf{42.2} & \textbf{15.9} & \textbf{22.5} \\
\cdashline{1-8}
Intern2.5-VL \citep{internvl-2.5}     & 8B  & 26.7 & 27.9 & 30.1 & 34.8 & 5.5 & 9.8 \\
Idefics2 \citep{idefics2}        & 8B  & 18.3 & 18.9 & 20.2 & 25.2 & 3.4 & 6.1 \\
\rowcolor{gray!30} \textbf{Idefics2\(_{\text{{\datasetname}}}\)} & 8B & 54.1 & 54.9 & 31.2 & 33.9 & 12.3 & 20.2 \\
\cdashline{1-8}
Qwen2.5-VL \citep{qwen2-5-vl} & 72B & 38.0 & 39.4 & 30.1 & 33.9 & 9.4 & 12.3 \\
\bottomrule
\end{tabular}}
}
\label{tab:{\datasetname}_refined_cot_result}
\end{table}

%% file: tables/cross_modal_benchmark_results.tex
\begin{table*}[t]
\centering
\caption{\textbf{Performance on Cross-Modal Reasoning Benchmarks.}
Performance comparison of zero-shot Idefics2-8B and the same model fine-tuned on Mantis-Instruct alone or jointly with {\datasetname}, where Mantis-Instruct and {\datasetname} have a 5:1 dataset size ratio.
\textit{\# Samples} indicates the total number of training samples in the corresponding fine-tuned dataset.
}
\setlength{\tabcolsep}{7pt}
\scalebox{\tablescale}{
\begin{tabular}{l c ccccc cc c c}
\toprule
 
&& \multicolumn{4}{c}{\textbf{SPIQA}}
& \multicolumn{2}{c}{\textbf{VEGA}}
& \multicolumn{2}{c}{\textbf{MMQA}}
& \textbf{FCMR} \\
\cmidrule(lr){3-6} \cmidrule(lr){7-8} \cmidrule(lr){9-10}

\textbf{Fine-tuned Dataset} & \textbf{\# Samples}
 & METEOR & ROUGE-L & CIDEr & BERTScore F1
 & ROUGE-L & BLEU
 & EM & F1
 & F1 \\
\midrule

\textit{No Fine-tuning} & -
 & 1.13 & 1.10 & 1.03 & 28.23
 & 31.4 & 6.7
 & 27.4 & 31.7
 & 40.5 \\

Mantis-Instruct & 837k
 & 3.60 & 10.17 & 23.83 & 42.83
 & 29.5 & 5.1
 & 27.3 & 30.9
 & 44.9 \\

Mantis-Instruct+{\datasetname} & 1005k
 & \textbf{10.53} & \textbf{22.77} & \textbf{67.93} & \textbf{55.80}
 & \textbf{35.1} & \textbf{7.1}
 & \textbf{30.0} & \textbf{33.8}
 & \textbf{50.5} \\

\bottomrule
\end{tabular}
}
\label{tab:cross_modal_benchmarks}
\end{table*}

%% file: tables/standard_multimodal_benchmark_result.tex
\begin{table*}[t]
\centering
\caption{\textbf{Performance on Standard Multimodal Benchmarks.}
Performance comparison of zero-shot Idefics2-8B and the same model fine-tuned on Mantis-Instruct alone or jointly with {\datasetname}, where Mantis-Instruct and {\datasetname} have a 5:1 dataset size ratio.
\textit{\# Samples} indicates the total number of training samples in the corresponding fine-tuned dataset.
}
\setlength{\tabcolsep}{7pt}
\scalebox{\tablescale}{
\begin{tabular}{l c c c c c c c c}
\toprule
\textbf{Fine-tuned Dataset}
& \textbf{\# Samples}
& \textbf{MuirBench}
& \textbf{BLINK}
& \textbf{MP-DocVQA}
& \textbf{MMStar}
& \textbf{MME}
& \textbf{SeedBench}
& \textbf{ChartQA} \\
\midrule

\textit{No Fine-tuning}
& -
& 26.2 & 45.2 & 46.7 & 30.2 & 1792 & 26.4 & 26.4 \\

Mantis-Instruct
& 837k
& \textbf{33.1} & 45.7 & 48.2 & 47.2 & 1909 & \textbf{67.3} & 63.8 \\

Mantis-Instruct+{\datasetname}
& 1005k
& 32.6 & \textbf{47.6} & \textbf{49.8} & \textbf{48.3} & \textbf{1910} & \textbf{67.3} & \textbf{64.9} \\

\bottomrule
\end{tabular}
}
\label{tab:mm_benchmark_comparison}
\end{table*}

%% file: tables/results_data_scaling.tex
\begin{table}[t]
\centering
\caption{\textbf{Results on {\datasetname} Extension.} 
Performance comparison of zero-shot Qwen2.5-VL-7B and the same model fine-tuned on the original {\datasetname} training set and an extended version augmented with additional samples. \textit{\# Samples} indicates the total number of training samples in the corresponding fine-tuned dataset.}
\scalebox{\tablescale}{%
\setlength{\tabcolsep}{4pt}
\begin{tabular}{lccccccc}
\toprule
 & \multicolumn{1}{c}{} 
 & \multicolumn{2}{c}{\textbf{NI}} 
 & \multicolumn{2}{c}{\textbf{VF}} 
 & \multicolumn{2}{c}{\textbf{SP}} \\
\cmidrule(lr){3-4} \cmidrule(lr){5-6} \cmidrule(lr){7-8}
\textbf{Fine-tuned Dataset} & \textbf{\# Samples} & EM & F1 & EM & F1 & EM & F1 \\
\midrule
\textit{No Fine-tuning} & -- & 28.3 & 29.1 & 24.0 & 27.8 & 6.8 & 9.6 \\
{\datasetname} & 84k & 58.6 & 59.5 & 38.8 & 42.2 & 15.9 & 22.5 \\
{\datasetname} Augmented & 210K & \textbf{62.6} & \textbf{63.7} & \textbf{45.6} & \textbf{49.1} & \textbf{16.7} & \textbf{22.8} \\
\bottomrule
\end{tabular}%
}
\label{tab:data_scaling}
\end{table}

%% file: sec/6_error_analysis.tex
\section{Error Analysis}

\begin{figure}[t]
    \centering
    \includegraphics[width=0.45\textwidth]{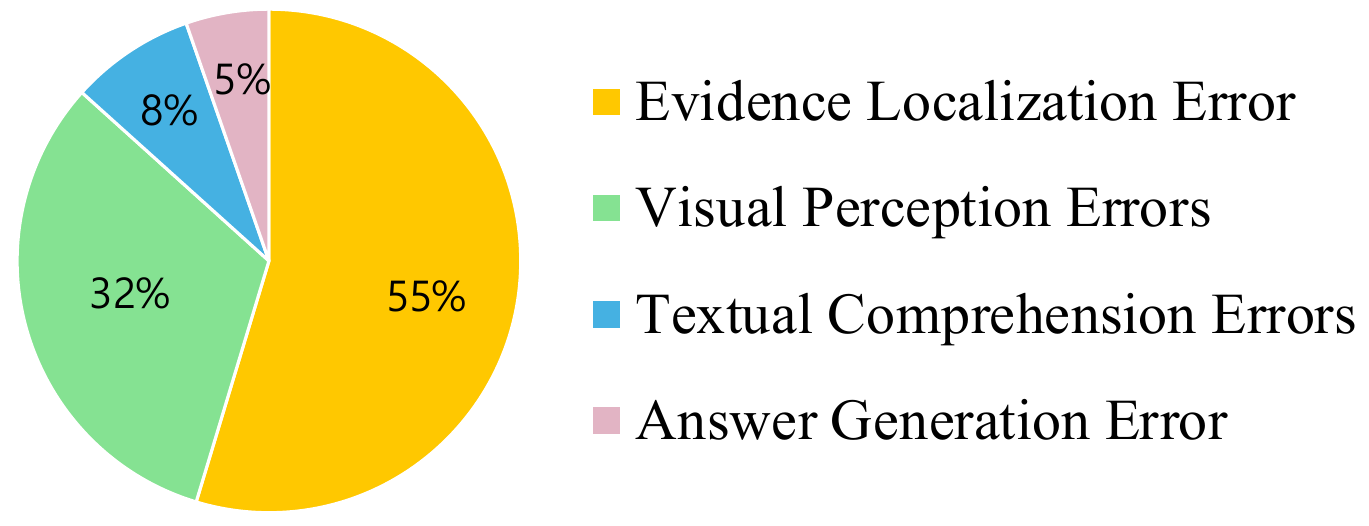}
        % \caption{\textbf{Error categories and their distribution across 75 manually annotated error cases for GPT-4o.}
        \caption{\textbf{Error categories and their distribution across 75 randomly sampled GPT-4o responses on CRIT.}}
    \label{fig:error_analysis}
\end{figure}
We conducted an error analysis to identify the key bottlenecks of current VLMs in cross-modal multi-hop reasoning tasks. Specifically, we use 75 randomly selected incorrect predictions from GPT-4o, each of which is manually verified to ensure the error is genuine before analysis. We categorize these failures into four types: 1) \textbf{Evidence Localization Error}: The model fails to correctly locate or ground the relevant evidence needed to answer the question. 2) \textbf{Visual Perception Error}: The model correctly identifies the relevant image but hallucinates or misperceives its visual content. 3) \textbf{Textual Comprehension Error}: The model hallucinates or misinterprets information within the textual context. 4) \textbf{Answer Generation Error}: The model correctly gathers relevant information but produces an incorrect final answer due to flawed reasoning. Figure~\ref{fig:error_analysis} shows the distribution of these errors. 
Among these, evidence localization error is the most prevalent, accounting for 55\% of all observed failures. These cases involve the model selecting the wrong image or text paragraph, or seeking information in the wrong modality. This highlights that our benchmark effectively reveals the limitations of current VLMs in cross-modal multi-hop reasoning, where models must integrate and reason over information from both visual and textual sources.
The second most frequent error type is visual perception error, which occurs roughly four times more often than textual comprehension error, suggesting that models remain comparatively robust in language understanding. 
The least common category is answer generation error, implying that when all relevant evidence is correctly identified, models are generally capable of producing the correct final answer.
Qualitative examples of each error type are provided in \supp

%% file: sec/7_conclusion.tex
\section{Conclusion}
We explored the challenge of cross-modal multi-hop reasoning, where solving a task requires weaving together complementary evidence from interleaved text and images. To advance this capability, we introduced {\datasetname}, built through a novel graph-based automatic pipeline that yields scalable, high-quality training and evaluation data. 
Fine-tuning on {\datasetname} delivers notable gains across diverse benchmarks without degrading their existing capabilities, and our augmentation experiments demonstrate a practical path to further scaling. We believe {\datasetname} provides a foundation for developing models with more robust and grounded multimodal reasoning.

\section*{Acknowledgements}
This work was supported by the IITP grants (RS-2025-02653113, IITP-2026-RS-2020-II201819, IITP-2026-RS-2024-00436857, RS-2024-00398115, IITP-2026-RS-2025-02304828), the NRF grant (RS-2025-23523979) and the KOCCA grant (RS-2024-00345025) funded by the Korean government (MSIT, MSCT).
% This work was supported by Institute of Information \& communications Technology Planning \& Evaluation (IITP) grant funded by the Korea government(MSIT) (RS-2025-02653113, High-Performance Research AI Computing Infrastructure Support at the 2 PFLOPS Scale).
% 생성형 ai 저쟉권 과제 (세환)
% 명품인재 (도성)
% itrc (리경)
% 생성형 과제 (수민)
% 스타펠로우쉽 (영서/진성)
% 연구재단 (영서/진성)

%% file: sec/X_suppl.tex
\clearpage
\setcounter{page}{1}
\maketitlesupplementary
\appendix

\section{Limitations}
Despite the strong performance gains achieved by models fine-tuned with CRIT—constructed via a graph-based automatic data generation pipeline—several limitations persist. First, the dataset is entirely synthesized by the LLM. Although the graph-based formulation imposes structural constraints and leverages human-annotated sources as initialization sources, the resulting QA pairs and reasoning chains may still reflect biases or stylistic patterns inherent to the generator model. In addition, the LLM-generated textual contexts may contain fictional or non-grounded content. That said, when evaluated on existing cross-modal reasoning benchmarks derived from diverse human-authored sources (e.g., SPIQA, VEGA, MMQA, FCMR), models trained with CRIT consistently demonstrate performance gains, suggesting that the learned capabilities generalize beyond the synthetic data distribution. Nevertheless, such evaluations may not fully surface subtle generator-specific biases, which could still be present but remain unobserved under current benchmarks. Second, the multi-stage nature of the data generation pipeline introduces potential error accumulation, as LLMs are employed throughout the process. Imperfections at earlier stages may propagate and amplify in subsequent steps. Finally, the pipeline incurs non-trivial computational overhead, which may pose challenges for efficiency and resource usage.

\section{Comparison with Prior Work}
Table \ref{tab:prior_work_comparison} summarizes the key differences between CRIT and existing datasets based on cross-modal reasoning. Although datasets such as SPIQA \cite{pramanick2024spiqa} and VEGA \cite{vega} adopt interleaved image-text inputs, and others like MMQA \cite{multimodalqa} and WebQA \cite{webqa} can be reformulated into this format, the majority of their questions are still solvable using a single modality. As shown in Table \ref{tab:single_modality_answerable}, this limits their ability to probe genuine cross-modal reasoning.
{\datasetname} is explicitly designed to overcome this limitation. Every question requires multi-hop, cross-modal inference, compelling models to integrate complementary visual and textual evidence rather than rely on isolated cues.

A second differentiator is domain diversity. Existing datasets commonly rely on narrow sources, such as Wikipedia~\citep{multimodalqa,webqa,manymodalqa} or specialized scientific and financial corpora~\citep{pramanick2024spiqa,vega,fcmr,mme-finance}. In contrast, CRIT draws from a broader and more heterogeneous set of public data, enabling evaluation of general-purpose multimodal understanding. Finally, CRIT emphasizes scalability and interpretability. Using the LLM-driven pipeline with manual verification of the test split, we construct 271k high-quality QA pairs. Our graph-based generation process additionally provides the exact subgraph used to form each question, yielding transparent reasoning traces that support training, diagnostics, and deeper analysis of model behavior.

\input{tables/single_modality_answerable}

\input{figures/img_domain}
\input{figures/text_domain}
\input{tables/comparison_with_prior_work}

\section{Domain Distribution of CRIT}
Figures \ref{fig:img_domain} and \ref{fig:text_domain} illustrate the distribution of image and text data across domains in the CRIT benchmark, demonstrating broad coverage with samples spanning diverse image categories and text genres.
\section{Data Generation Detail for Video and Text-Rich Sources}
This section provides further details on (1) how video captions are used to form a unified graph across sampled frames, (2) how text-rich sources are converted into multimodal graphs, and (3) how textual context is refined when paragraphs refer to visual elements.

For video sources, the full set of captions along with their temporal ranges is provided to the LLM. The model identifies all entities mentioned throughout the video, extracts their attributes, and performs coreference resolution to merge entities appearing across multiple frames. Frame-level relations are then inferred, capturing interactions and events described in each segment. Compared with natural images, the resulting video graphs are more action-centric and reflect the temporal structure of the content. The prompt used for caption-to-graph conversion is shown in Figure~\ref{fig:prompt_caption_to_graph}.

Scientific papers are processed directly from their raw TeX sources. After segmenting the text into paragraphs, we prompt the LLM to extract entities and textual relations for paragraphs that do not reference any figures or tables, using the prompt shown in Figure~\ref{fig:prompt_paper_nonfig_paragraph_to_graph}. For paragraphs that do contain such references, we extract only relations grounded in the referenced visual content—e.g., numerical results (“Model A achieves 92\% accuracy”), comparative statements (“Method B outperforms Method C”), or visual attributes (“The blue curve increases sharply”), following the prompt in Figure~\ref{fig:prompt_paper_fig_paragraph_to_graph}.

To ensure that questions require genuine cross-modal reasoning, paragraphs referencing figures or tables undergo sentence-level filtering. After visual relations are extracted, we embed each sentence in the paragraph using SBERT (all-MiniLM-L6-v2)~\citep{sbert} and compute cosine similarity with the extracted visual relations. Sentences that closely overlap with the visual content are removed according to a fixed threshold, while all remaining sentences are retained and used as the textual context. Paragraphs without visual references are left unchanged. This yields textual context that complements rather than duplicates the visual evidence, promoting reasoning chains that span modalities.

\section{Failure Cases of Data Generation Without Graphs}
Figures \ref{fig:llm_generation_failure_case_1} and \ref{fig:llm_generation_failure_case_2} illustrate failure cases that arise when prompting the LLM to directly produce cross-modal multi-hop QA pairs from raw image–text inputs without any additional structural guidance.
For each example, one image and its associated paragraph are randomly sampled from the interleaved image-text context, and the LLM is asked to produce a QA pair requiring cross-modal multi-hop reasoning given the caption of the sampled image and the paragraph.
Figure~\ref{fig:llm_generation_failure_case_1} shows a case where the generated question appears cross-modal but is actually solvable using a single modality. The model implicitly incorporates visual information into the question itself, making the textual paragraph alone sufficient for answering. Figure~\ref{fig:llm_generation_failure_case_2} shows a question that is syntactically framed as multi-hop, yet relies on only a single reasoning step.
These cases underscore typical shortcomings observed when prompting LLMs without additional structural constraints, motivating the graph-based approach introduced in the main paper.
\input{tables/results_on_others_including_spiqa_ablation}

\section{Ablation Study on Data Generation Pipeline}
We conduct a retrospective ablation study to assess the contribution of key components in our data generation pipeline. Experiments are performed on 100 randomly sampled instances from the CRIT test set, using GPT-5~\citep{gpt5} as an automatic evaluator.
We first evaluate the role of cross-image edges in the multimodal content graph. These edges connect entities across different images via textual nodes and enable multi-hop reasoning that spans multiple images. When removing all cross-image edges, we find that $33\%$ of the questions become unanswerable. This demonstrates that such connections are essential for maintaining valid cross-modal reasoning paths.
Next, we examine the impact of graph-based QA generation. We compare QA pairs generated using our graph-structured pipeline with those produced via direct prompting without explicit graph guidance. Using GPT-5 as a judge with the prompt shown in Figure~\ref{fig:prompt_judge_compare}, graph-based QA pairs are preferred in $77\%$ of cases. This preference indicates that graph-based generation produces questions that are more coherent, better grounded in the underlying multimodal content, and more consistent with the intended multi-hop reasoning chains.
Overall, the results highlight the importance of both cross-image connectivity and graph-based generation in producing high-quality cross-modal multi-hop reasoning data.

\section{Benchmarks and Evaluation}
This section provides additional details on the benchmarks and clarifies how evaluation scores are reported.

\noindent\textbf{SPIQA} \  \ 
SPIQA~\citep{pramanick2024spiqa} is an open-ended QA dataset tailored for interpreting complex figures and tables in scientific research articles across diverse subfields of computer science. Each instance contains interleaved image–text inputs, where figures and tables are provided as images accompanied by their captions. The test set comprises three splits. We report the average performance across them.

\noindent\textbf{VEGA} \  \
VEGA~\citep{vega} is a dataset designed for interleaved image-text comprehension, which requires models to identify relevant visual and textual regions within a complex multimodal context and generate accurate answers to the given questions. The test set includes two splits differing in input token length. We report averaged results over both.

\noindent\textbf{MMQA} \  \
MMQA~\citep{multimodalqa} is a multi-source question-answering benchmark constructed from Wikipedia, requiring reasoning over text passages, tables, and images. The dataset includes both single-hop and multi-hop questions, with some cases demanding cross-modal integration to infer the correct answer. In our setting, we provide all text, image, and table inputs at once in an interleaved format. We report multi-hop performance on the validation set.

\noindent\textbf{FCMR} \  \
FCMR \cite{fcmr} evaluates multimodal reasoning in the financial domain, combining textual reports, tables, and charts. Each instance provides three candidate statements whose verification requires evidence aggregation across modalities. The official all-or-nothing scoring scheme counts a prediction as correct only when the exact set of true statements is returned; in practice, this criterion is overly strict and leads models to behave nearly randomly. To provide a more informative assessment, we instead report per-statement F1 scores computed using the ground-truth labels, treating each statement as an independent binary decision so that partial correctness is properly reflected.

\noindent\textbf{MuirBench} \  \
MuirBench \cite{muirbench} comprises 12 diverse multi-image understanding tasks formulated as multiple-choice questions. We follow the evaluation protocol provided by LMMs-Eval~\citep{lmms-eval}.

\noindent\textbf{BLINK} \  \
BLINK \cite{BLINK} targets core perceptual skills that humans can typically solve “within a blink”, including depth ordering, correspondence, forensics detection, and multi-view reasoning. Several tasks involve multiple images, and all are posed as multiple-choice questions. We report performance on the validation split using VLMEvalKit~\citep{vlmevalkit}.

\noindent\textbf{MP-DocVQA} \  \
MP-DocVQA \cite{mp-docvqa} is a multi-page document QA benchmark requiring models to parse text, layout, and visual elements across pages to locate the relevant page and produce an answer. We report performance on the validation split using LMMs-Eval~\citep{lmms-eval}.

% \noindent\textbf{MMMStar} \  \
% We follow the evaluation protocol provided by LMMs-Eval~\citep{lmms-eval}.

% \noindent\textbf{MME} \  \
% We follow the evaluation protocol provided by LMMs-Eval~\citep{lmms-eval}.

% \noindent\textbf{SeedBench} \  \
% We follow the evaluation protocol provided by LMMs-Eval~\citep{lmms-eval}.

% \noindent\textbf{ChartQA} \  \
% We report performance on the test split using LMMs-Eval~\citep{lmms-eval}.
\noindent\textbf{MMStar} \  \
MMStar~\citep{mmstar} is a multimodal benchmark designed to evaluate VLMs under strictly visual-dependent settings. It focuses on samples that require genuine visual understanding, reducing cases where questions can be answered without images. We follow the evaluation protocol provided by LMMs-Eval~\citep{lmms-eval}.

\noindent\textbf{MME} \  \
MME~\citep{mme} is a comprehensive benchmark for assessing both perception and cognition abilities of multimodal models. It consists of multiple subtasks covering object recognition, OCR, commonsense reasoning, and knowledge-based understanding. We follow the evaluation protocol provided by LMMs-Eval~\citep{lmms-eval}.

\noindent\textbf{SeedBench} \  \
SeedBench~\citep{seedbench} is a large-scale multimodal benchmark that evaluates hierarchical vision-language capabilities through multiple-choice questions. It covers a wide range of tasks, including visual understanding, spatial reasoning, and multimodal comprehension. We follow the evaluation protocol provided by LMMs-Eval~\citep{lmms-eval}.

\noindent\textbf{ChartQA} \  \
ChartQA~\citep{chartqa} is a benchmark for question answering over charts, requiring models to interpret visual elements such as bars, lines, and legends, and perform reasoning over chart data. We report performance on the test split using LMMs-Eval~\citep{lmms-eval}.

\section{Ablation Study on Fine-Tuning Data Composition}
%To verify that the performance gains reported in Table~\ref{tab:other_benchmarks} stem from incorporating {\datasetname}, we also jointly train on Mantis-Instruct and SPIQA sampled to match the size of {\datasetname} in place of {\datasetname}. 
To verify that the performance gains observed when jointly fine-tuning on Mantis-Instruct and {\datasetname} truly come from incorporating {\datasetname}, we also fine-tune on Mantis-Instruct combined with a size-matched subset of SPIQA in place of {\datasetname}. Aside from this substitution, all training settings and data quantities are kept identical to ensure a fair comparison.
While SPIQA is also an interleaved image–text dataset, it contains single-hop, open-ended QA pairs, and therefore does not provide the multi-hop reasoning supervision present in {\datasetname}. 
%Aside from this substitution, all training settings and data quantities are kept identical to ensure a fair comparison.

The results in Table~\ref{tab:other_benchmarks_ablation} show that jointly training with SPIQA instead of {\datasetname} leads to consistently lower performance across nearly all benchmarks. The performance drop is particularly pronounced on tasks that require multi-hop or cross-modal reasoning (e.g., VEGA, MMQA, FCMR). Surprisingly, fine-tuning with {\datasetname} also achieves stronger results on several SPIQA evaluation metrics—METEOR, ROUGE-L, and BERTScore-F1—despite SPIQA being the source of those benchmarks. Overall, these findings confirm that the improvements observed when jointly fine-tuning Mantis-Instruct with {\datasetname} are indeed attributable to incorporating {\datasetname}.

\input{tables/train_hyperparemeter}
\input{tables/result_on_crit_reasoning_models}

\section{Training Hyperparameters}
Training was conducted using 4 NVIDIA H100 80GB GPUs. The training hyperparameters used for these models are summarized in Table \ref{tab:hyperparams}.

\section{Performance on {\datasetname} with Reasoning Models}
Table~\ref{tab:crit_result_reasoning_models} summarizes the performance of VLMs specialized for reasoning, including Intern3-VL~\citep{internvl3}, Qwen-3-VL-Thinking~\citep{qwenvl3}, GLM-4.1V-Thinking~\citep{glmv}, and Kimi-VL-Thinking~\citep{kimi-vl}. Although reasoning-oriented models generally outperform their non-reasoning counterparts, Qwen2.5-VL$_{\text{{\datasetname}}}$ achieves the best performance across all domains, despite having the smallest model size.

\section{Performance on {\datasetname} with Direct Answer}
Table \ref{tab:refined_da_result} reports the performance of several VLMs on the CRIT benchmark under the \textit{direct answer} setting, where models are prompted to produce final answers without any intermediate reasoning. Consistent with the trends observed in the CoT setting, both Qwen2.5-VL$_\text{{\datasetname}}$ and Idefics2$_\text{{\datasetname}}$ yield substantial gains over their respective baselines across all domains.

\section{Performance on {\datasetname} by Hops}
Table \ref{tab:result_by_hop} summarizes performance by hop count over all domains of {\datasetname}. Somewhat unexpectedly, performance does not consistently decline as the required hop count increases. A key reason is that even when the annotated hop count is higher, the underlying visual reasoning demanded by the question often does not become more complex. Instead, the additional steps typically involve textual reasoning, which current models handle relatively well. Models are generally strong at processing the textual components of the task, so increases in textual reasoning depth do not substantially affect accuracy. Rather, performance is more dependent on how effectively a model can locate and extract the relevant information within an image—such as recognizing objects, interpreting spatial relationships, or identifying fine-grained visual cues. Errors therefore tend to arise from visual information retrieval failures rather than from the number of reasoning steps required. For future work, we plan to expand the evaluation set and conduct a more extensive analysis to better understand the underlying factors that drive these patterns.

\input{tables/results_on_crux_refined_da}
\input{tables/result_on_crux_by_hop}

\section{Qualitative Examples of Error Analysis}
We conducted a systematic error analysis of GPT-4o’s performance on {\datasetname} to investigate its limitations in cross-modal multi-hop reasoning. By manually reviewing 75 incorrect responses, we identified four distinct error categories. Representative examples and detailed analysis of these categories are shown in Figure~\ref{fig:visual_perception_error} through ~\ref{fig:answer_generation_error}.

\section{Details of Human Verification}
\label{appendix:human_verification}

\subsection{Guideline for Task}
For each sample, raters are provided with one or more \textbf{images} (such as natural photographs, video frames, or figures from research papers), along with a \textbf{textual context}. They also receive a \textbf{QA pair}, as well as the \textbf{subgraph} used to generate it.
The task of the rater is to judge whether the given QA pair constitutes a valid example of cross-modal multi-hop reasoning. Each sample must be classified into one of three categories:
\begin{itemize}
  \item \textbf{Keep}: Valid example that should be included in the benchmark.
  \item \textbf{Discard}: Invalid example that should not be included in the benchmark.
  \item \textbf{Unsure}: Uncertain case that requires further review or discussion.
\end{itemize}

\subsection{Checklist for Decision-Making}
To ensure consistency, the raters receive a checklist to decide whether to \textbf{Keep} or \textbf{Discard} a sample.

\paragraph{Keep the sample if:}
\begin{itemize}
  \item The question can only be answered by integrating \textbf{both image(s) and text}.
  \item The reasoning requires \textbf{multi-hop inference} (chaining across multiple entities, edges, or modalities).
  \item The answer is correct, unambiguous, and consistent with the given evidence.
  \item The question is natural, clear, and does not contain awkward phrasing or hallucinations.
  \item There is only one answer and no other alternatives that fit the evidence.
\end{itemize}

\paragraph{Discard the sample if:}
\begin{itemize}
  \item The question can be answered using  \textbf{only text} or  \textbf{only image(s)} without cross-modal reasoning.
  \item The reasoning does not actually require  \textbf{multi-hop} (i.e., a single entity lookup is enough).
  \item The answer is incorrect, incomplete, or contradictory.
  \item The question is  \textbf{ill-posed} (ambiguous, nonsensical, ungrammatical, or explicitly revealing the reasoning trace).
  \item The question ignores information from the subgraph or incorrectly conveys relationships.
  \item There are multiple valid answers or the answer is subjective.
\end{itemize}

\subsection{Verification Result}
12 English-proficient annotators with STEM backgrounds participated in the verification process, and retained 41\% of the samples. 
Most discarded samples were not incorrect, but were excluded because they had multiple valid answers or were subjective, which were strictly filtered for evaluation reliability. 
The average inter-annotator agreement on overlapping subsets is 90.5\%.

\section{Qualitative Examples of CRIT Benchmark}
We provide additional examples from the CRIT benchmark in Figures~\ref{fig:crux_example_1} and \ref{fig:crux_example_2}.

\section{Prompts for Data Generation}

\subsection{Graph Construction Prompt}
We employ six types of Text Node Generation Prompts. Below, we present three representative examples. In addition, we include the Edge Generation Prompt.
See Figure~\ref{fig:prompt_graph_construction} for an example prompt.

\subsection{Textual Context Generation Prompt}
For the Textual Context Generation Prompt, we define a set of diverse context types. These include Story/Narrative, Newspaper Article, Comedy Sketch, Diary Entry, Poem, Song Lyrics, Documentary Script, Blog Post, Motivational Speech, Promotional Article, Movie Scene Description, and Social Media Post. We randomly use one of them in each prompt.
See Figure~\ref{fig:prompt_textual_context_generation} for an example prompt.

\subsection{Question Answer Pair Generation Prompt}
See Figure~\ref{fig:prompt_question_answer_pair_generation} for an example prompt.

\subsection{CoT Response Generation Prompt}
See Figure~\ref{fig:prompt_cot_response_generation} for an example prompt.

\subsection{Caption-to-Graph Conversion Prompt}
See Figure~\ref{fig:prompt_caption_to_graph} for an example prompt.

\subsection{Graph Transformation for Paragraph Not Containing Figure Reference Prompt}
See Figure~\ref{fig:prompt_paper_nonfig_paragraph_to_graph} for an example prompt.

\subsection{Graph Transformation for Paragraph Containing Figure Reference Prompt}
See Figure~\ref{fig:prompt_paper_fig_paragraph_to_graph} for an example prompt.

\input{figures/llm_data_generation_failure_1}
\input{figures/llm_data_generation_failure_2}

\input{figures/visual_perception_error}
\input{figures/evidence_localization_error}
\input{figures/textual_comprehension_error}
\input{figures/answer_generation_error}

\input{figures/crux_examples}

\input{figures/prompt_w_wo_graph_compare}
\input{figures/prompt_graph_construction}

\input{figures/prompt_textual_context_generation}

\input{figures/prompt_question_answer_pair_generation}

\input{figures/prompt_cot_response_generation}

\input{figures/prompt_caption_to_graph}

\input{figures/prompt_paper_nonfig_paragraph_to_graph}

\input{figures/prompt_paper_fig_paragraph_to_graph}

%% file: tables/single_modality_answerable.tex
% \begin{table}[t]
% \centering
% \caption{\textbf{Results of Single Modality Input.}
% Performance comparison of zero-shot Qwen2.5-VL-7B on benchmarks consisting of interleaved image-text format. 
% We compare input settings using image only, text only, or both.}
% \label{tab:single_modality_answerable}

% \scalebox{\tablescale}{%
% \setlength{\tabcolsep}{4pt}
% \begin{tabular}{cccccc}
% \toprule
% \textbf{Image} & \textbf{Text} & \textbf{MMQA} & \textbf{SPIQA} & \textbf{VEGA} & \textbf{CRIT} \\
% \midrule
% \cmark & \cmark & -- & -- & -- & -- \\ % Image+Text
% \cmark & \xmark & -- & -- & -- & -- \\ % Image only
% \xmark & \cmark & -- & -- & -- & -- \\ % Text only
% \bottomrule
% \end{tabular}
% }
% \end{table}

%%%%
\begin{table}[t]
\centering
\caption{\textbf{Ablation on Image Input.}
We evaluate GPT-4o on benchmarks formulated as interleaved image–text inputs with and without images. Performance remains strong on existing benchmarks but drops sharply on {\datasetname}, indicating that {\datasetname} is designed to require cross-modal reasoning. M: METEOR, R-L: ROUGE-L, C: CIDEr, B-F1: BERTScore F1, B: BLEU.}
\label{tab:single_modality_answerable}
\setlength{\tabcolsep}{4pt}
\scalebox{\tablescale}{
\begin{tabular}{c cccc cc cc cc}
\toprule
 
& \multicolumn{4}{c}{\textbf{SPIQA}}
& \multicolumn{2}{c}{\textbf{VEGA}}
& \multicolumn{2}{c}{\textbf{MMQA}}
& \multicolumn{2}{c}{\textbf{CRIT}} \\
\cmidrule(lr){2-5} \cmidrule(lr){6-7} \cmidrule(lr){8-9} \cmidrule(lr){10-11}

\textbf{Image}
 & M & R-L & C & B-F1
 & R-L & B
 & EM & F1
 & EM & F1 \\
\midrule

% Image + Text
\cmark & 17.3 & 30.8 & 88.0 & 57.8 & 40.1 & 12.3 & 48.8 & 56.0 & 28.0 & 32.3 \\

% Text only
\xmark & 11.8 & 19 & 36.6 & 49.9 & 40.3 & 12.7 & 42.9 & 49.6 & 10.5 & 12.5 \\

\bottomrule
\end{tabular}
}
\end{table}

%% file: figures/img_domain.tex
\begin{figure}[t]
\centering
\includegraphics[width=1\columnwidth]
{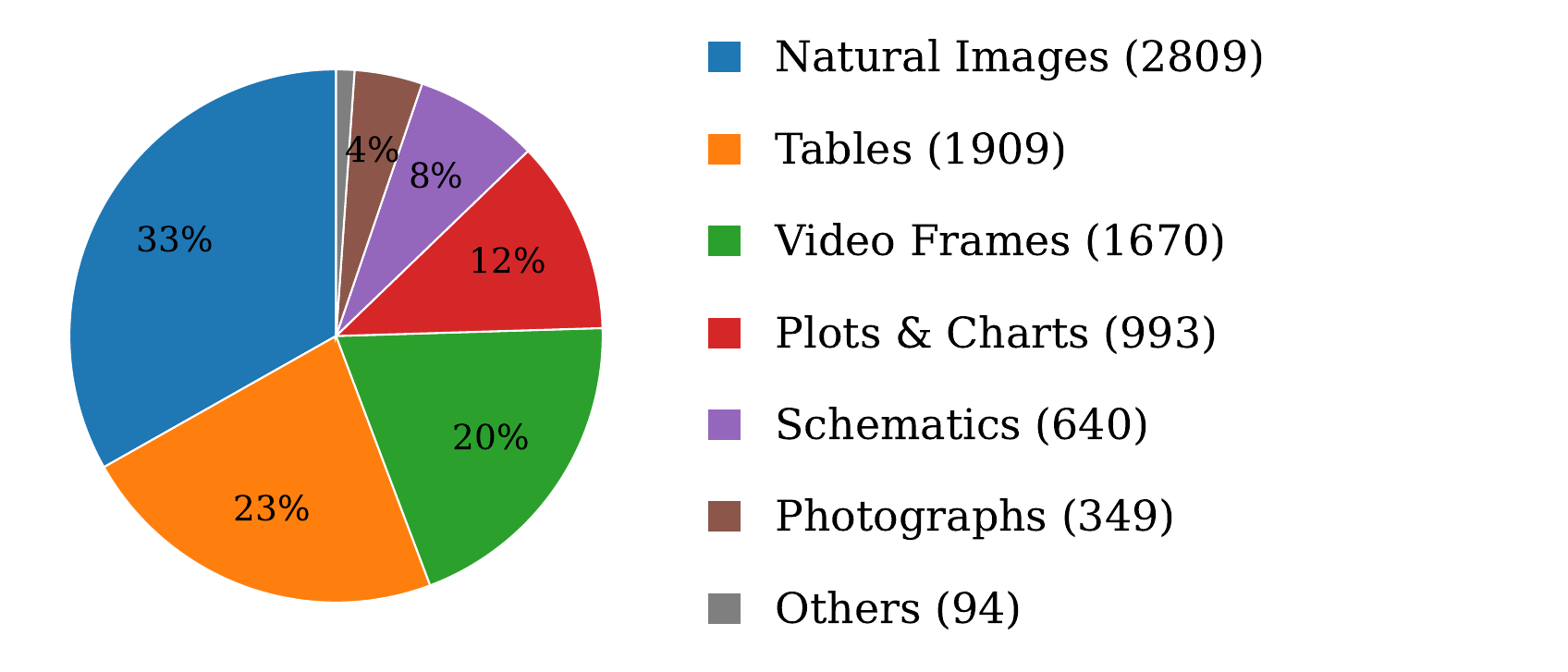}
\caption{\textbf{Image Distribution}}
\label{fig:img_domain}
\end{figure}

%% file: figures/text_domain.tex
\begin{figure}[t]
\centering
\includegraphics[width=1\columnwidth]
{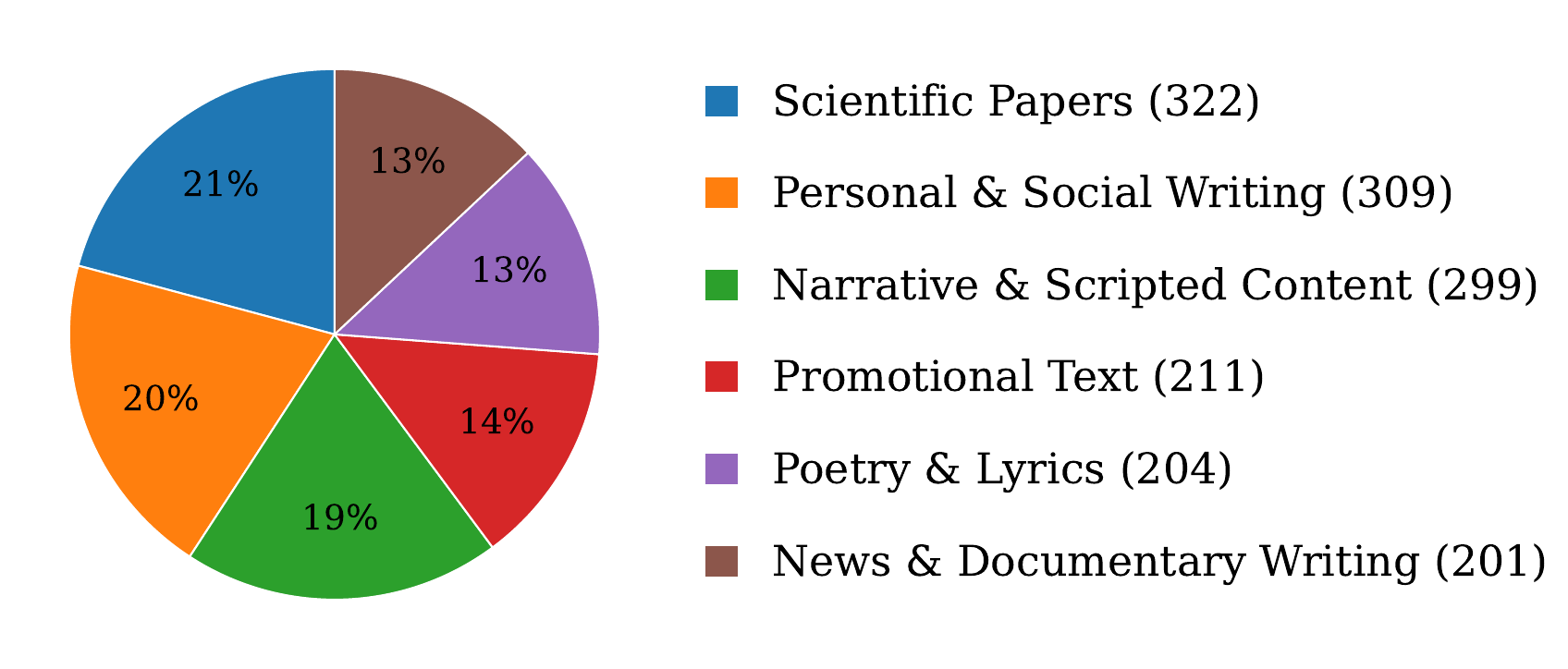}
\caption{\textbf{Text Distribution}}
\label{fig:text_domain}
\end{figure}

%% file: tables/comparison_with_prior_work.tex
% \begin{table*}[t]
% \centering
% \caption{Comparison of our dataset with prior related works. 
% \textbf{Annotation:} \cmark: Human Annotation, \xmark: Automatic Annotation, \cxmark: Semi-automatic Annotation.}
% \scalebox{\tablescale}{%
% %\setlength{\tabcolsep}{5pt}
% \begin{tabular}{lcccccc}
% \hline
% \textbf{Dataset} &
% \makecell{\textbf{Interleaved}\\\textbf{Image-Text}} &
% \makecell{\textbf{Multi}\\\textbf{Hop}} &
% \makecell{\textbf{Diverse}\\\textbf{Domain}} &
% \makecell{\textbf{Data}\\\textbf{Generation}} &
% \makecell{\textbf{Reasoning}\\\textbf{Trace}} &
% \makecell{\textbf{Num}\\\textbf{QA}} \\
% \hline
% ManyModalQA~\citep{manymodalqa} & \xmark & \xmark & - & - & - & -- \\
% CT2C-QA~\citep{ct2c}            & \xmark & \xmark & - & - & - & -- \\
% MME-Finance~\citep{mme-finance} & \xmark & \xmark & - & - & - & -- \\
% WebQA~\citep{webqa}             & \xmark & \cmark & - & - & - & -- \\
% MuMuQA~\citep{mumuqa}           & \xmark & \cmark & - & - & - & -- \\
% MMQA~\citep{multimodalqa}       & \xmark & \cmark & - & - & - & -- \\
% FCMR~\citep{fcmr}               & \xmark & \cmark & - & - & - & -- \\
% SPIQA~\citep{pramanick2024spiqa} & \cmark & \xmark & - & - & - & -- \\
% VEGA~\citep{vega}               & \cmark & \xmark & - & - & - & -- \\
% \hline
% \textbf{Ours}                   & \cmark & \cmark & \cmark & \cmark & \cmark & -- \\
% \hline
% \end{tabular}
% }
% \label{tab:prior_work_comparison}
% \end{table*}

\begin{table*}[t]
\centering
\caption{\textbf{Comparison of our dataset with prior cross-modal reasoning works.}
{\datasetname} targets cross-modal multi-hop reasoning over interleaved image-text inputs, constructed through a graph-based automatic pipeline spanning diverse data sources, and provides 271k QA pairs together with explicit reasoning traces.
}
\scalebox{\tablescale}{%
\begin{tabular}{lcccccc}
\hline
\textbf{Dataset} &
\textbf{Interleaved Image-Text} &
\textbf{Multi Hop} &
\textbf{Diverse Data Source} &
\textbf{Data Generation} &
\textbf{Reasoning Trace} &
\textbf{Num QA} \\
\hline
ManyModalQA~\citep{manymodalqa} & \xmark & \xmark & \xmark & Human experts & \xmark & 10.1k \\
CT2C-QA~\citep{ct2c}            & \xmark & \xmark & \xmark & LLMs + Human experts & \xmark & 10k \\
MME-Finance~\citep{mme-finance} & \xmark & \xmark & \xmark & LLMs + Human experts & \xmark & 2.3k \\
WebQA~\citep{webqa}             & \xmark & \cmark & \xmark & Human experts & \xmark & 46.7k \\
MuMuQA~\citep{mumuqa}           & \xmark & \cmark & \xmark & Human experts & \xmark & 1.4k \\
MMQA~\citep{multimodalqa}       & \xmark & \cmark & \xmark & Template + Human experts & \xmark & 30k \\
FCMR~\citep{fcmr}               & \xmark & \cmark & \xmark & LLMs + Human experts & \xmark & 2.2k \\
SPIQA~\citep{pramanick2024spiqa} & \cmark & \xmark & \xmark & LLMs + Human experts & \cmark & 270k \\ 
VEGA~\citep{vega}               & \cmark & \xmark & \xmark & LLMs + Human experts & \xmark & 286k \\
\hline
\textbf{\datasetname}                   & \cmark & \cmark & \cmark & LLMs + Human experts & \cmark & 271k \\
\hline
\end{tabular}
}
\label{tab:prior_work_comparison}
\end{table*}

%% file: tables/results_on_others_including_spiqa_ablation.tex
\begin{table*}[t]
\centering
\caption{\textbf{Ablation of Fine-Tuning Data Composition}
Performance comparison of zero-shot Idefics2-8B and the same model fine-tuned on different datasets. We compare fine-tuning on Mantis-Instruct alone and jointly with CRIT or SPIQA. Mantis-Instruct and CRIT follow a 5:1 sampling ratio, and all joint settings contain the same total number of training samples. Joint training with CRIT provides the strongest overall improvements, while size-matched SPIQA yields notably smaller gains, underscoring the contribution of {\datasetname}.
M: METEOR, R-L: ROUGE-L, C: CIDEr, B-F1: BERTScore F1, B: BLEU, Acc: Accuracy.
}
\setlength{\tabcolsep}{5pt}
\scalebox{\tablescale}{
\begin{tabular}{l c ccccc cc c c c c c}
\toprule
 
&& \multicolumn{4}{c}{\textbf{SPIQA}}
& \multicolumn{2}{c}{\textbf{VEGA}}
& \multicolumn{2}{c}{\textbf{MMQA}}
& \textbf{FCMR} 
& \textbf{MuirBench} 
& \textbf{BLINK} 
& \textbf{MP-DocVQA} \\
\cmidrule(lr){3-6} \cmidrule(lr){7-8} \cmidrule(lr){9-10}

\textbf{Fine-tuned Dataset} & \textbf{\# Samples}
 & M & R-L & C & B-F1
 & R-L & B
 & EM & F1
 & F1 & Acc & Acc & Acc \\
\midrule

\textit{No Fine-tuning} & -
 & 1.13 & 1.10 & 1.03 & 28.23
 & 31.4 & 6.7
 & 27.4 & 31.7
 & 40.5 & 26.2 & 45.2 & 46.7 \\

%\cdashline{1-14}
Mantis-Instruct & 837k
 & 3.60 & 10.17 & 23.83 & 42.83
 & 29.5 & 5.1
 & 27.3 & 30.9
 & 44.9 & 33.1 & 45.7 & 48.2 \\

 % Mantis-Instruct subset+{\datasetname} & 837k
 % & 6.07 & 17.27 & 45.97 & 50.73
 % & 28.6 & 3.0
 % & 29.1 & 33.7
 % & 63.4 & 31.0 & \textbf{47.8} & \textbf{50.0} \\

%\cdashline{1-14}
Mantis-Instruct+{\datasetname} & 1005k
 & \textbf{10.53} & \textbf{22.77} & 67.93 & \textbf{55.80}
 & \textbf{35.1} & \textbf{7.1}
 & \textbf{30.0} & \textbf{33.8}
 & \textbf{50.5} & 32.6 & \textbf{47.6} & \textbf{49.8} \\

 Mantis-Instruct+SPIQA & 1005k
 & 10.0 & 22.47 & \textbf{71.7} & 54.57
 & 28.4 & 4.0
 & 29.2 & 32.3
 & 47.5 & \textbf{34.9} & 46.4 & 48.8 \\

\bottomrule
\end{tabular}
}
\label{tab:other_benchmarks_ablation}
\end{table*}

%% file: tables/train_hyperparemeter.tex
\begin{table}[t]
\centering
\caption{\textbf{Training hyperparameters for Qwen2.5-VL and Idefics2.}}
\label{tab:hyperparams}
\scalebox{\tablescale}{
\begin{tabular}{lcc}
\hline
\textbf{Hyperparameter} & \textbf{Qwen2.5-VL\textsubscript{CRIT}} & \textbf{Idefics2\textsubscript{CRIT}} \\
\hline
Epochs         & 1    & 1 \\
Batch size     & 64   & 128 \\
Learning rate  & 2e-5 & 2e-5 \\
Optimizer      & AdamW & AdamW \\
Warmup ratio   & 0.05 & 0.05 \\
Scheduler      & cosine & cosine \\
LoRA rank      & 32   & 128 \\
LoRA alpha     & 64   & 256 \\
LoRA dropout   & 0.05  & 0.1 \\
\hline
\end{tabular}
}
\end{table}

%% file: tables/result_on_crit_reasoning_models.tex
\begin{table}[htbp]
\centering
\caption{\textbf{Results on {\datasetname} including reasoning models.}
Performance comparison of reasoning models and fine-tuned VLM across natural image (NI), video frame (VF), and scientific paper (SP).
Models fine-tuned on {\datasetname} are highlighted with a gray background.
\textit{\# Prm} denotes the number of model parameters.
}
%Performance comparison of proprietary, open-source, and fine-tuned VLMs across natural image (NI), video frame (VF), and scientific paper (SP).
%Models fine-tuned on {\datasetname} are highlighted with a gray background.
%Open-source models are grouped by model size for clearer comparison.
%\textit{\# Prm} denotes the number of model parameters.}
\setlength{\tabcolsep}{5pt} % Reduce horizontal padding between columns
\scalebox{\tablescale}{{\begin{tabular}{lccc ccc ccc}
\toprule
 & 
& \multicolumn{2}{c}{\textbf{NI}} 
& \multicolumn{2}{c}{\textbf{VF}} 
& \multicolumn{2}{c}{\textbf{SP}} \\
\cmidrule(lr){3-4} \cmidrule(lr){5-6} \cmidrule(lr){7-8}
\textbf{Model}&\textbf{\# Prm} & EM & F1 & EM & F1 & EM & F1 \\
\midrule

%\multicolumn{8}{c}{\textbf{\textit{Proprietary Models}}} \\
%\midrule
%GPT-4o      & - & 35.1 & 37.7 & 32.0 & 38.9 & 8.4 & 14.0 \\
%\midrule
\rowcolor{gray!30} \textbf{Qwen2.5-VL\(_{\text{{\datasetname}}}\)} & 7B & \textbf{58.6} & \textbf{59.5} & \textbf{38.8} & \textbf{42.2} & \textbf{15.9} & \textbf{22.5} \\
%\cdashline{1-8}
Intern3-VL~\citep{internvl3}   & 8B  & 33.1 & 34.3 & 35.2 & 41.3 & 7.3 & 12.1 \\
%\cdashline{1-8}
Qwen3-VL-Thinking~\citep{qwenvl3}   & 8B  & 31.8 & 32.9 & 33.1 & 37.1 & 7.3 & 9.6 \\
%\cdashline{1-8}
GLM-4.1V-Thinking~\citep{glmv} & 9B  & 12.0 & 17.7 & 11.2 & 12.1 & 5.2 & 10.4 \\
%\cdashline{1-8}
Kimi-VL-Thinking~\citep{kimi-vl}   & 16B  & 33.8 & 35.3 & 33.9 & 38.1 & 8.1 & 12.3 \\
%\cdashline{1-8}
Qwen3-VL-Thinking~\cite{qwenvl3} & 32B  & 43.1 & 44.2 & 37.2 & 41.9 & 8.4 & 10.5 \\
%\cdashline{1-8}
\bottomrule
\end{tabular}}
}
\label{tab:crit_result_reasoning_models}
\end{table}

%% file: tables/results_on_crux_refined_da.tex
\begin{table}[t]
\centering
\caption{\textbf{Results on {\datasetname} with Direct Answer.}
Performance comparison of proprietary, open-source, and fine-tuned VLMs across natural image (NI), video frame (VF), and scientific paper (SP).
Models fine-tuned on {\datasetname} are highlighted with a gray background.
Open-source models are grouped by model size for clearer comparison.
\textit{\# Prm} denotes the number of model parameters.}
\label{tab:refined_da_result}
\setlength{\tabcolsep}{5pt} % Reduce horizontal padding between columns
\scalebox{\tablescale}{{\begin{tabular}{lccc ccc ccc}
\toprule
 & 
& \multicolumn{2}{c}{\textbf{NI}} 
& \multicolumn{2}{c}{\textbf{VF}} 
& \multicolumn{2}{c}{\textbf{SP}} \\
\cmidrule(lr){3-4} \cmidrule(lr){5-6} \cmidrule(lr){7-8}
\textbf{Model}&\textbf{\#Prm} & EM & F1 & EM & F1 & EM & F1 \\
\midrule

\multicolumn{8}{c}{\textbf{\textit{Proprietary Models}}} \\
\midrule
GPT-4o \citep{gpt-4o}       & - & 21.0 & 21.4 & 32.8 & 39.2 & 10.2 & 13.0 \\
GPT-4o-mini \citep{gpt-4o}   & - & 24.9 & 26.0 & 31.7 & 36.0 & 9.4 & 13.3 \\
\midrule

\multicolumn{8}{c}{\textbf{\textit{Open-Source Models}}} \\
\midrule
Phi3.5-Vision \citep{phi}    & 4B  & 24.1 & 25.4 & 29.5 & 31.8 & 5.5 & 9.5 \\
\cdashline{1-8}
LLaVA-Onevision \citep{llava_onevision} & 7B  & 28.9 & 29.7 & 33.3 & 36.8 & 6.3 & 10.1 \\
Qwen2.5-VL \citep{qwen2-5-vl}       & 7B  & 27.7 & 28.9 & 35.0 & 38.5 & 11.0 & 14.2 \\
\rowcolor{gray!30} \textbf{Qwen2.5-VL\(_{\text{{\datasetname}}}\)} & 7B & \textbf{61.0} & \textbf{61.8} & \textbf{38.8} & \textbf{42.6} & 14.4 & 20.8 \\
\cdashline{1-8}
Intern2.5-VL \citep{internvl-2.5}     & 8B  & 37.3 & 37.9 & 33.9 & 38.4 & 7.3 & 11.0 \\
Idefics2 \citep{idefics2}        & 8B  & 15.7 & 16.5 & 28.4 & 31.8 & 2.9 & 5.4 \\
\rowcolor{gray!30} \textbf{Idefics2\(_{\text{{\datasetname}}}\)} & 8B & 54.9 & 55.7 & 34.4 & 36.8 & \textbf{15.4} & \textbf{21.4} \\
\cdashline{1-8}
Qwen2.5-VL \citep{qwen2-5-vl} & 72B & 38.0 & 39.2 & 36.9 & 40.3 & 10.2 & 14.6 \\

\bottomrule
\end{tabular}}
}
\end{table}

%% file: tables/result_on_crux_by_hop.tex
\begin{table*}[t]
\centering
\caption{\textbf{Results on {\datasetname} by Hops.}
Performance comparison of proprietary, open-source, and fine-tuned VLMs across natural image.
Models fine-tuned on {\datasetname} are highlighted with a gray background.
Open-source models are grouped by model size for clearer comparison.
\textit{\# Prm} denotes the number of model parameters.}
\scalebox{\tablescale}{
\begin{tabular}{lcccc cccc cccc cccc}
\toprule
&
& \multicolumn{2}{c}{\textbf{Overall}}
& \multicolumn{2}{c}{\textbf{2-hop}}
& \multicolumn{2}{c}{\textbf{3-hop}}
& \multicolumn{2}{c}{\textbf{4-hop}}
& \multicolumn{2}{c}{\textbf{5-hop}} \\

\cmidrule(lr){3-4}
\cmidrule(lr){5-6}
\cmidrule(lr){7-8}
\cmidrule(lr){9-10}
\cmidrule(lr){11-12}

\textbf{Model} & \textbf{\# Prm}
& EM & F1
& EM & F1
& EM & F1
& EM & F1
& EM & F1 \\

\midrule
\multicolumn{12}{c}{\textbf{\textit{Proprietary Models}}} \\
\midrule

Gemini 2.0 Flash \citep{gemini} & - 
& 26.0 & 28.8
& 26.2 & 28.6
& 22.2 & 26.3
& 27.9 & 29.7
& 30.5 & 34.3 \\

GPT-4o \citep{gpt-4o} & -
& 28.0 & 32.3
& 28.4 & 32.2
& 25.1 & 31.1 
& 27.2 & 29.3 
& 32.5 & 38.2  \\

GPT-4o-mini \citep{gpt-4o} & -
& 22.0 & 24.3
& 22.0 & 23.4 
& 19.5 & 24.6
& 22.1 & 24.1
& 27.8 & 30.0  \\

\midrule
\multicolumn{12}{c}{\textbf{\textit{Open-Source Models}}} \\
\midrule

Phi3.5-Vision \citep{phi} & 4B
& 17.7 & 19.7 
& 19.2 & 20.9 
& 14.6 & 16.8 
& 15.4 & 18.2 
& 16.6 & 19.4  \\

\cdashline{1-12}

LLaVA-Onevision \citep{llava_onevision} & 7B
& 25.8 & 28.0
& 25.9 & 27.7
& 23.3 & 27.2
& 25.0 & 26.6
& 31.1 & 32.8 \\

Qwen2.5-VL \citep{qwen2-5-vl} & 7B
& 22.2 & 24.1 
& 22.2 & 23.6 
& 21.0 & 24.6 
& 19.1 & 20.3 
& 27.2 & 29.2 \\

\rowcolor{gray!30}
\textbf{Qwen2.5-VL\(_{\text{{\datasetname}}}\)} & 7B
& \textbf{43.9} & \textbf{46.7}
& 46.9 & 49.6
& 35.0 & 38.5 
& 39.0 & 42.3
& 49.0 & 51.1 \\

\cdashline{1-12}

Intern2.5-VL \citep{internvl-2.5} & 8B
& 22.4 & 25.1
& 22.2 & 24.5
& 20.1 & 25.0 
& 21.3 & 22.8 
& 29.8 & 31.8  \\

Idefics2 \citep{idefics2} & 8B
& 15.2 & 17.3
& 15.0 & 16.6 
& 13.4 & 16.0 
& 18.4 & 19.6 
& 17.9 & 20.6  \\

\rowcolor{gray!30}
\textbf{Idefics2\(_{\text{{\datasetname}}}\)} & 8B
& 38.9 & 41.8
& 42.4 & 44.8 
& 28.6 & 33.7
& 30.9 & 34.0
& 47.0 & 48.5 \\

\cdashline{1-12}

Qwen2.5-VL \citep{qwen2-5-vl} & 72B
& 29.4 & 31.7 
& 28.2 & 30.2
& 25.1 & 29.0
& 31.6 & 33.0 
& 44.4 & 45.7 \\

\bottomrule
\end{tabular}
}
\label{tab:result_by_hop}
\end{table*}

%% file: figures/llm_data_generation_failure_1.tex
\begin{figure*}[t]
\begin{tcolorbox}[
    title=Failure Case 1, 
    width=\textwidth, sharp corners=southwest, 
    left=1mm, right=1mm, top=0mm, bottom=0mm
]

\includegraphics[height=4cm, width=9cm]{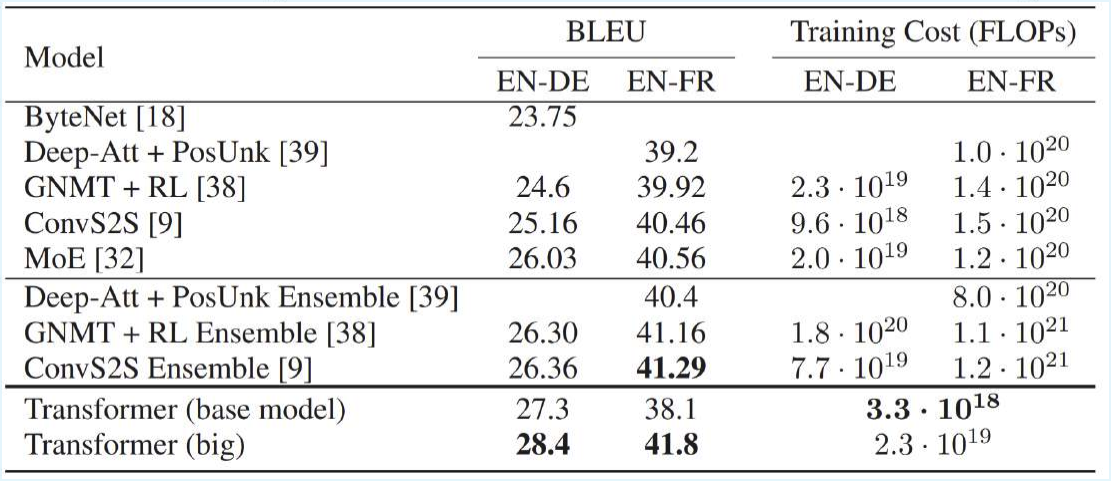}

\vspace{1em}

\textbf{Image Caption}: The Transformer achieves better BLEU scores than previous state-of-the-art models on the English-to-German and English-to-French newstest2014 tests at a fraction of the training cost.

\vspace{1em}

\textbf{Paragraph}: The goal of reducing sequential computation also forms the foundation of the Extended Neural GPU, ByteNet and ConvS2S, all of which use convolutional neural networks as basic building block, computing hidden representations in parallel for all input and output positions. In these models, the number of operations required to relate signals from two arbitrary input or output positions grows in the distance between positions, linearly for ConvS2S and logarithmically for ByteNet. This makes it more difficult to learn dependencies between distant positions. In the Transformer this is reduced to a constant number of operations, albeit at the cost of reduced effective resolution due to averaging attention-weighted positions, an effect we counteract with Multi-Head Attention as described in section 3.2. Self-attention, sometimes called intra-attention is an attention mechanism relating different positions of a single sequence in order to compute a representation of the sequence. Self-attention has been used successfully in a variety of tasks including reading comprehension, abstractive summarization, textual entailment and learning task-independent sentence representations. End-to-end memory networks are based on a recurrent attention mechanism instead of sequence-aligned recurrence and have been shown to perform well on simple-language question answering and language modeling tasks. To the best of our knowledge, however, the Transformer is the first transduction model relying entirely on self-attention to compute representations of its input and output without using sequence-aligned RNNs or convolution. In the following sections, we will describe the Transformer, motivate self-attention and discuss its advantages over models such as those previously mentioned.

\vspace{1em}

\textbf{Generated Question}: Based on the paragraph and the caption, why might the Transformer model achieve high BLEU scores despite having a potential limitation of reduced effective resolution due to averaging attention-weighted positions?

\vspace{1em}

\textbf{Generated Answer}: Because the Transformer uses Multi-Head Attention, which counteracts the reduced effective resolution caused by averaging attention-weighted positions.

\vspace{1em}

\textbf{Generated Reasoning Step}: The paragraph explains that while self-attention in the Transformer leads to reduced effective resolution due to averaging, this issue is mitigated by Multi-Head Attention, as described in section 3.2. The image confirms the Transformer's high BLEU scores indicating strong performance. Thus, combining the paragraph's explanation of the solution (Multi-Head Attention) with the image's evidence of high performance (high BLEU scores), we can infer that Multi-Head Attention enables the Transformer to overcome its limitation and achieve excellent results.

\vspace{1em}

\textbf{Analysis}: Although the question is presented as cross-modal, it already embeds the key information that should have been inferred from the image—namely, that the Transformer achieves high BLEU scores. Because this information is stated directly in the question itself, the reasoning no longer requires consulting the image. The remaining elements of the question can be fully answered using only the text paragraph, which explicitly explains how Multi-Head Attention mitigates the reduced effective resolution. As a result, the supposed cross-modal dependency collapses: the question is answerable from a single modality because it includes information that should be derived from the image.

\end{tcolorbox}
\caption{\textbf{Failure Case I: Generating Cross-Modal Multi-Hop QA with Direct Prompting}}
\label{fig:llm_generation_failure_case_1}
\end{figure*}

%% file: figures/llm_data_generation_failure_2.tex
\begin{figure*}[t]
\begin{tcolorbox}[
    title=Failure Case 2, 
    width=\textwidth, sharp corners=southwest, 
    left=1mm, right=1mm, top=0mm, bottom=0mm
]

\includegraphics[height=4cm, width=5cm]{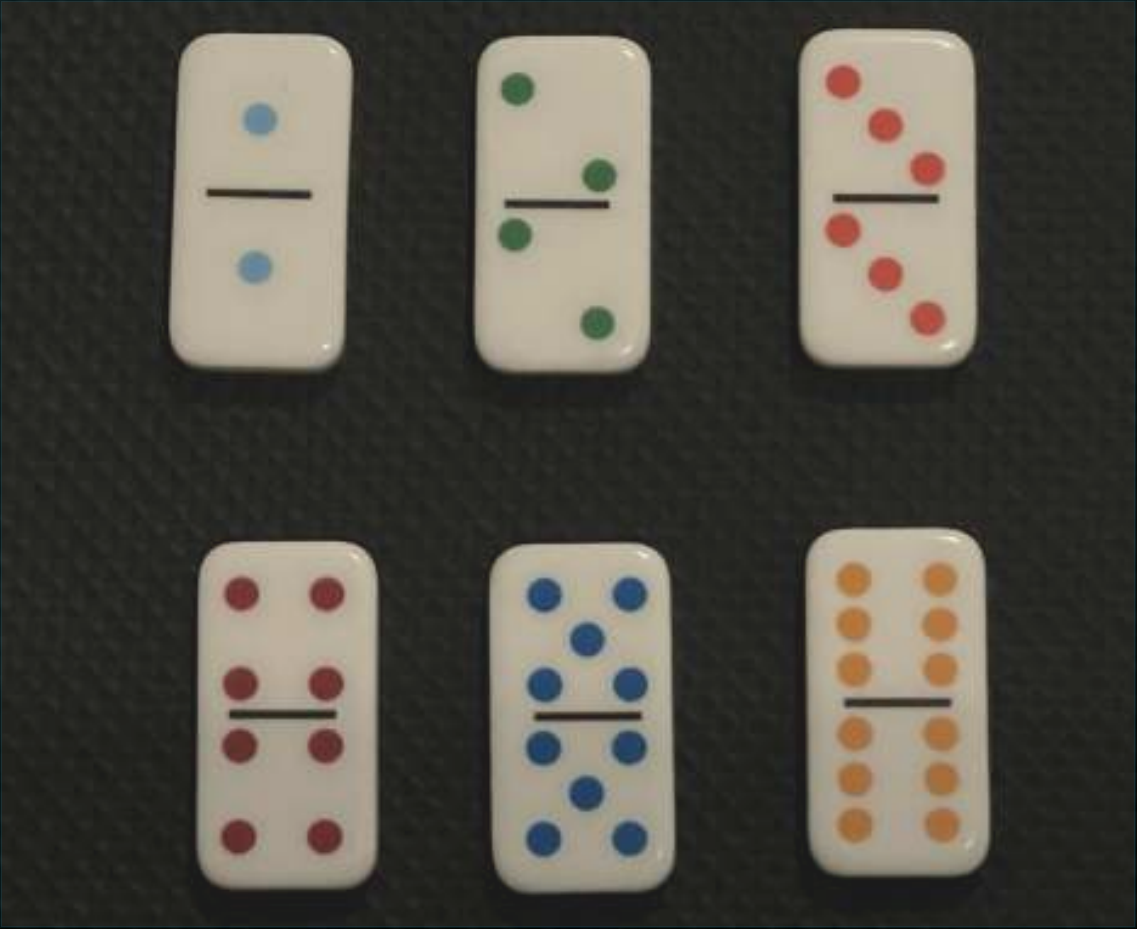}

\vspace{1em}

\textbf{Image Caption}: Top-down view of 6 beige dominoes with colored dots, long side vertically, in a 3x2 grid on a black diamond pattern plastic material. From left to right, top to bottom: light blue double 1, green double 2, light red double 3, dark red double 4, dark blue double 5, orange double 6.

\vspace{1em}

\textbf{Paragraph}: The old game room smelled of dust and forgotten afternoons. In the dim light filtering through the cracked blinds, a young girl named Mira knelt beside a worn wooden table, her fingers hovering over a peculiar arrangement of dominoes. She had found them tucked inside a dusty suitcase her grandmother had left behind, along with a note. The dominoes were arranged in a perfect rectangle, their surfaces cool and smooth, each one bearing a different color and pattern that seemed to shift under the flickering overhead bulb. Mira traced the edge of the one labeled with two light blue dots, whispering the number “one” as if it might awaken something. Suddenly, the room grew quiet, and the air hummed faintly, like a tuning fork struck in another dimension. She blinked, and for a moment, the black diamond-patterned surface beneath the dominoes seemed to ripple like water, revealing a glimpse of a bustling marketplace under a golden sky. Before she could react, the door creaked open. Her older brother, Theo, stepped inside, brushing cobwebs from his jacket. “You disappeared again,” he muttered, though the sharpness in his voice softened when he saw the suitcase. Behind him stumbled Samir, their neighbor and self-proclaimed explorer, clutching a half-broken flashlight as if it were a sacred artifact. “What did you find this time?” Samir asked, leaning over the table—only to jerk back when the green double two pulsed faintly beneath his shadow. The air grew colder. A soft voice—distant but unmistakably their grandmother’s—whispered, “Don’t turn them all at once.” Theo stiffened. “Did you hear that?” Mira nodded, her heart pounding. But before she could speak, a figure appeared at the edge of the room: Aunt Liora, the family’s enigmatic archivist, wrapped in a deep crimson scarf. No one had heard her enter. “So,” she murmured, eyes narrowing at the glowing dominoes, “it’s beginning again.” Mira hesitated, fingertips trembling above the pieces.

\vspace{1em}

\textbf{Generated Question}: Given that the dominoes are arranged in a 3x2 grid with their long sides vertical, and that the first domino Mira touches is the light blue double one, what is the likely sequence of dominoes she will encounter as she moves from left to right and top to bottom?

\vspace{1em}

\textbf{Generated Answer}: The sequence is light blue double one, green double two, light red double three, dark red double four, dark blue double five, orange double six.

\vspace{1em}

\textbf{Generated Reasoning Step}: The paragraph describes Mira touching the light blue double one domino first, which corresponds to the first domino in the grid. The image confirms a 3x2 grid with a top-to-bottom, left-to-right order: light blue double one, green double two, light red double three, dark red double four, dark blue double five, orange double six

\vspace{1em}

\textbf{Analysis}: The LLM was instructed to generate a three-hop question in which the first two hops rely on textual information and only the final hop requires using the image. However, the resulting question already embeds the key textual conclusions that should have been derived through the earlier hops—such as the rectangular arrangement, the 3×2 grid configuration, and the orientation of the dominoes. Because this information is stated outright, the intermediate textual reasoning becomes unnecessary. The model can immediately proceed to the image-derived ordering task without performing the intended multi-step progression. As a result, the question collapses the designed three-hop structure, demonstrating a failure to enforce genuine cross-modal multi-hop reasoning.

\end{tcolorbox}
\caption{\textbf{Failure Case II: Generating Cross-Modal Multi-Hop QA with Direct Prompting}}
\label{fig:llm_generation_failure_case_2}
\end{figure*}

%% file: figures/visual_perception_error.tex
\begin{figure*}[t]
\centering
\includegraphics[width=1\textwidth]{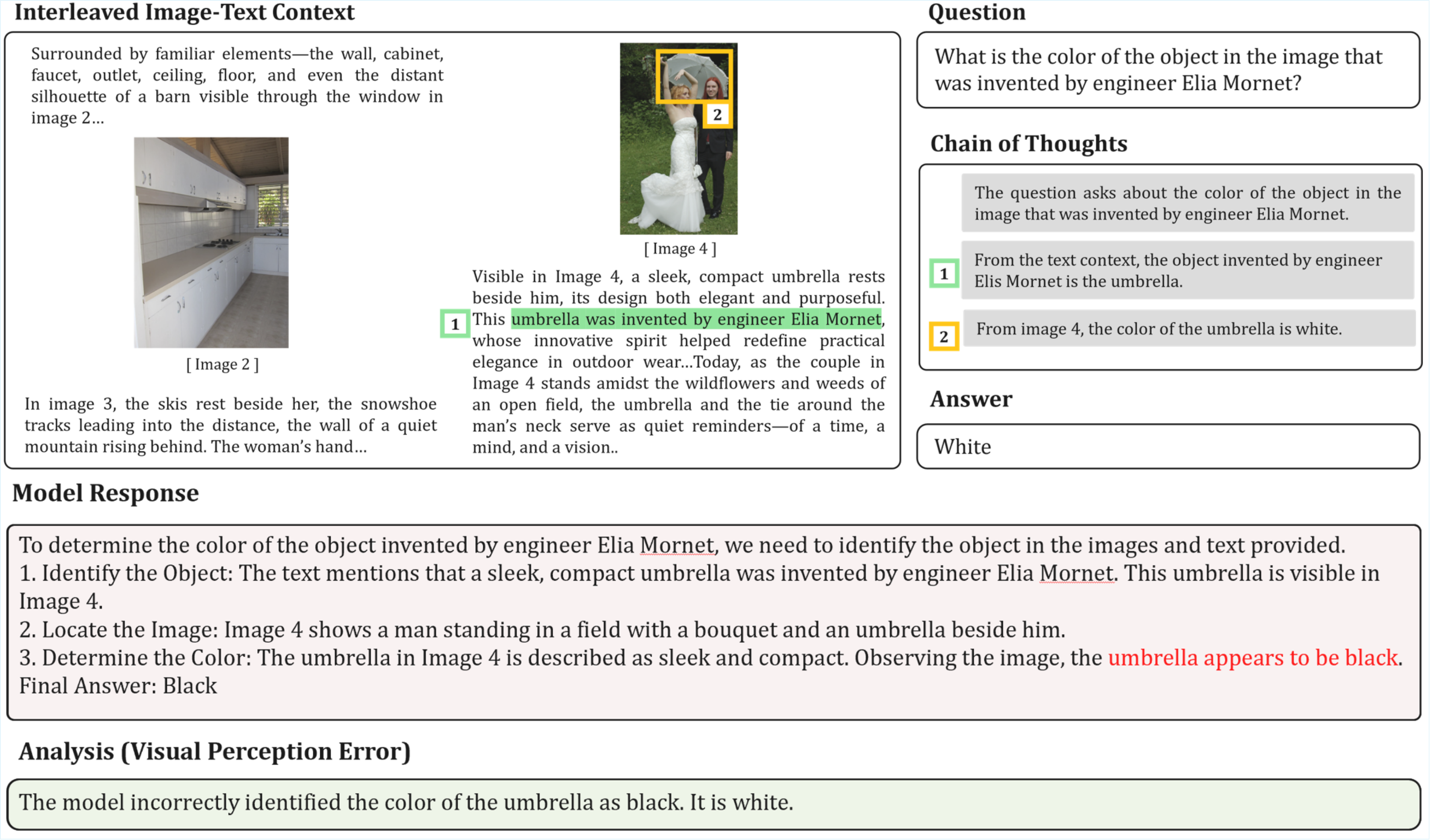}
\caption{\textbf{Illustration of Visual Perception Error Case.}}
\label{fig:visual_perception_error}
\end{figure*}

%% file: figures/evidence_localization_error.tex
\begin{figure*}[t]
\centering
\includegraphics[width=1\textwidth]{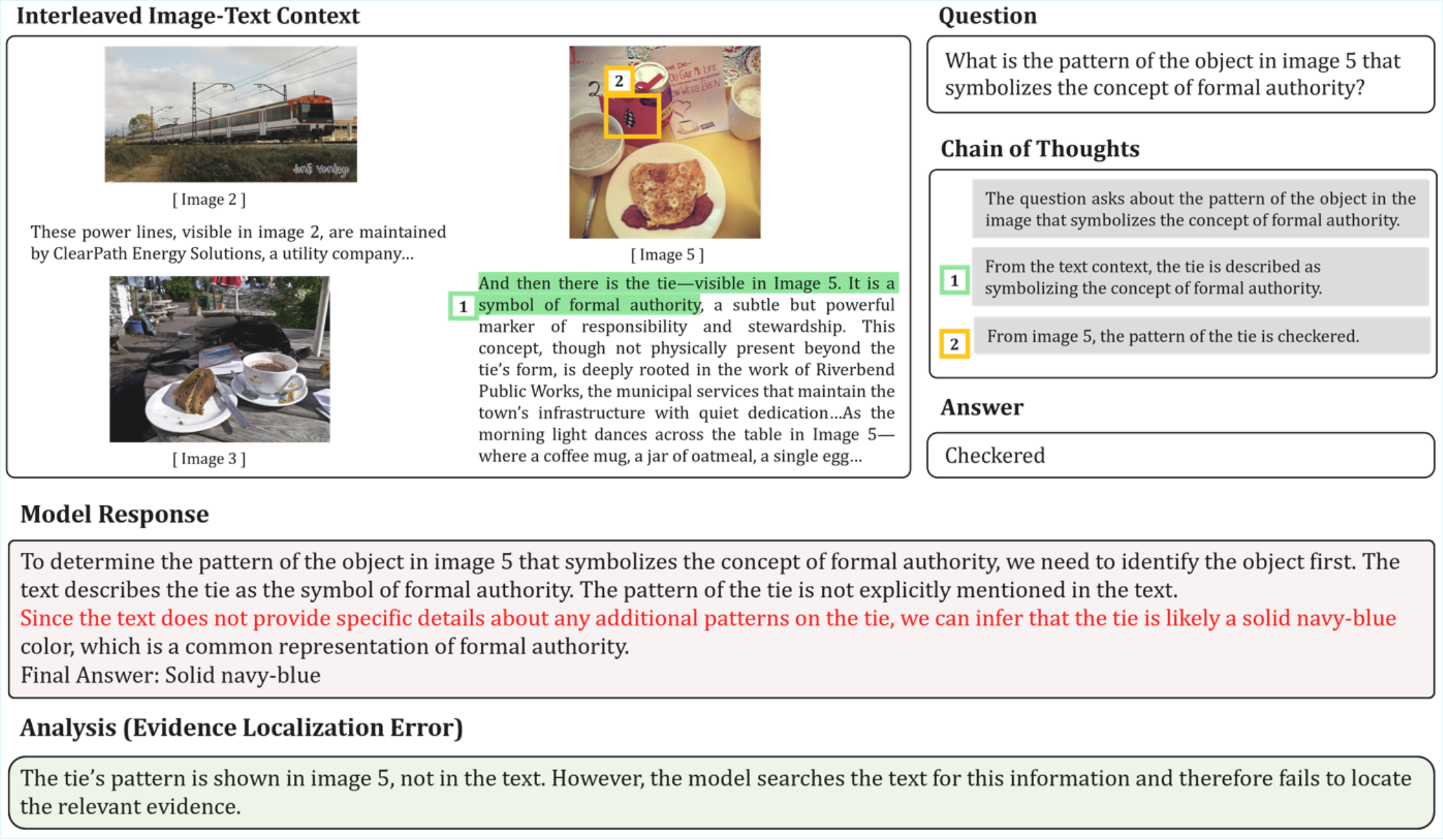}
\caption{\textbf{Illustration of Evidence Localization Error Case.}}
\label{fig:evidence_localization_error}
\end{figure*}

%% file: figures/textual_comprehension_error.tex
\begin{figure*}[t]
\centering
\includegraphics[width=1\textwidth]{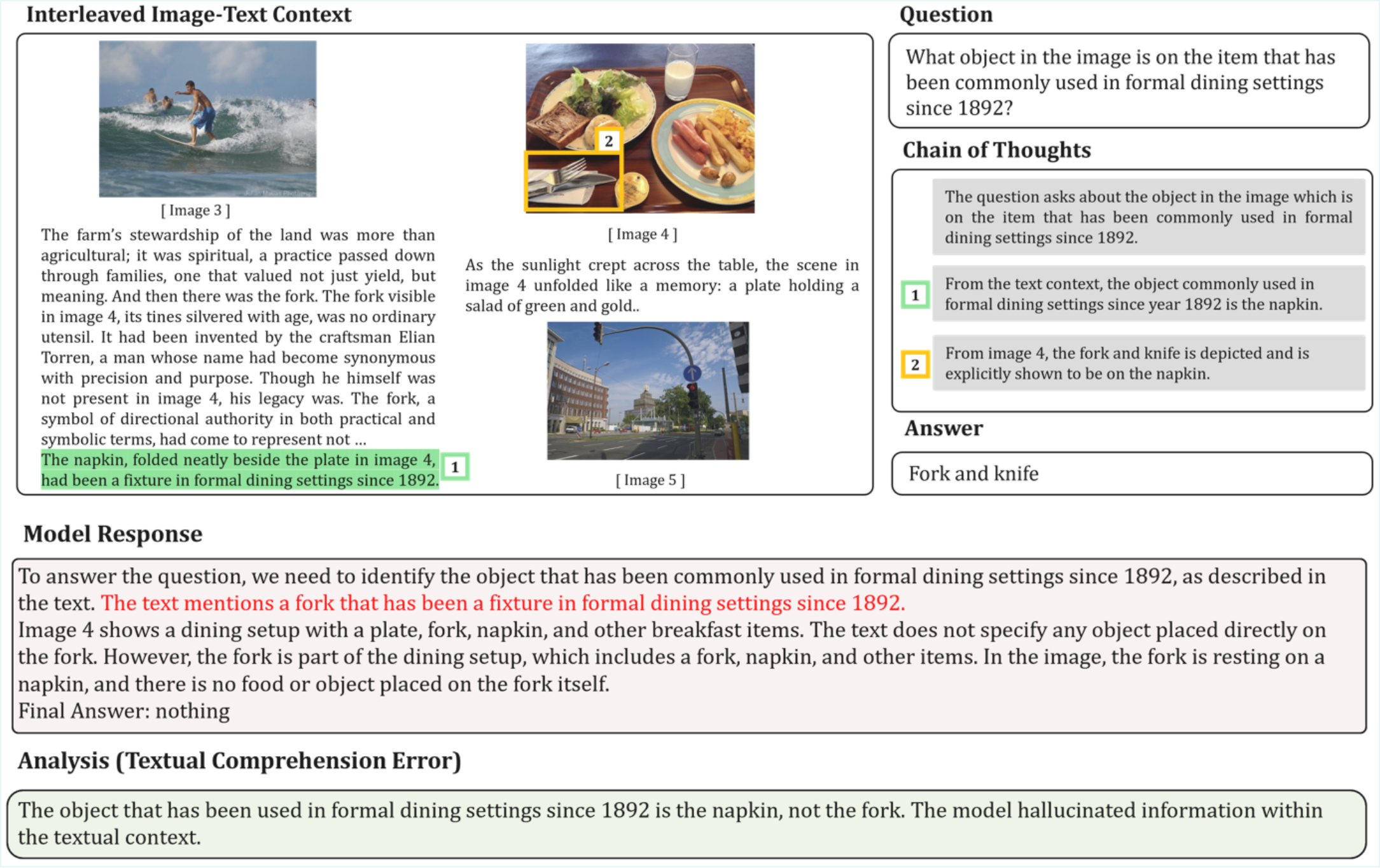}
\caption{\textbf{Illustration of Textual Comprehension Error Case.}}
\label{fig:textual_comprehension_error}
\end{figure*}

%% file: figures/answer_generation_error.tex
\begin{figure*}[t]
\centering
\includegraphics[width=1\textwidth]{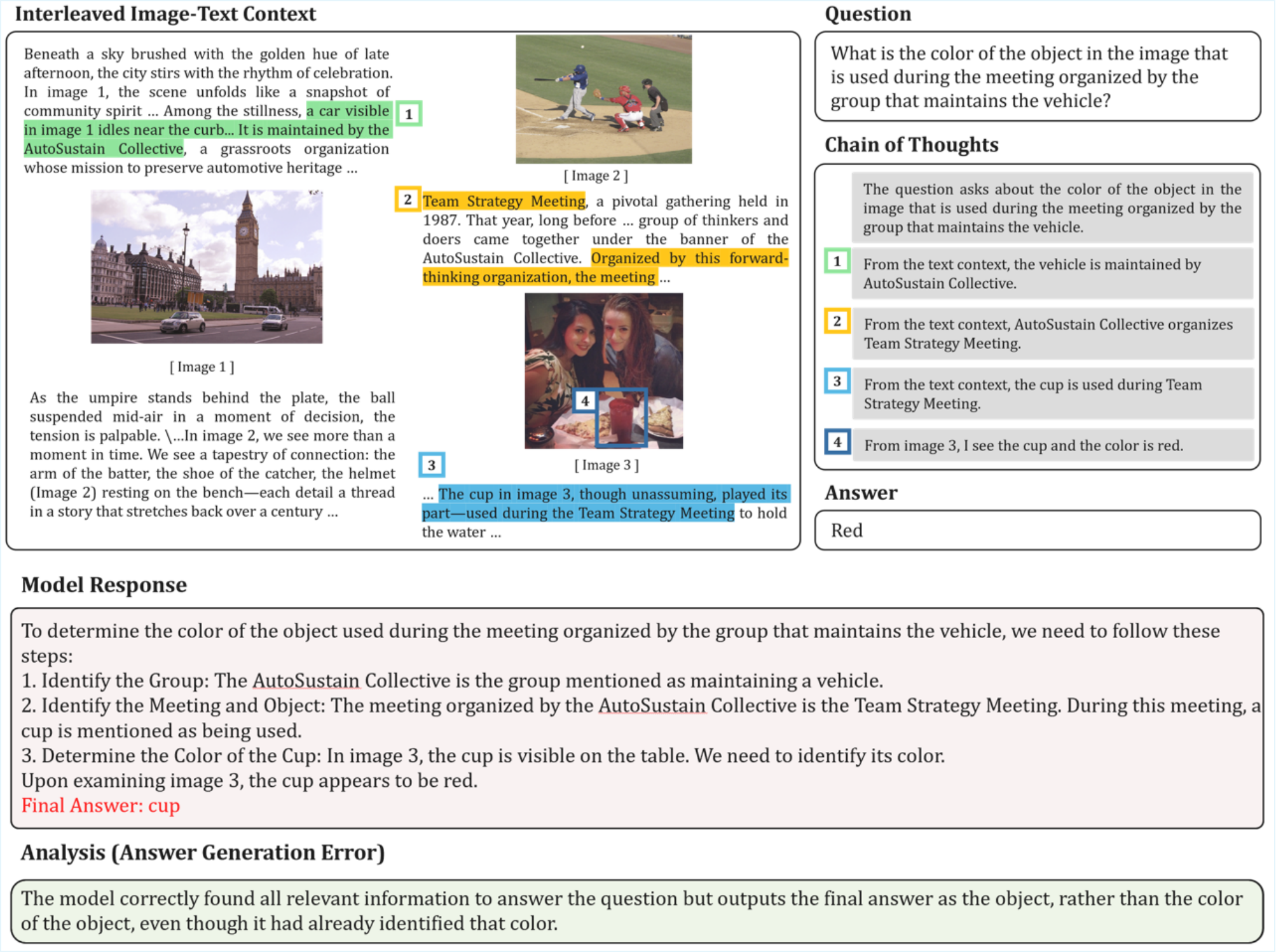}
\caption{\textbf{Illustration of Answer Generation Error Case.}}
\label{fig:answer_generation_error}
\end{figure*}

%% file: figures/crux_examples.tex
\begin{figure*}[t]
\centering
\includegraphics[width=1\textwidth]{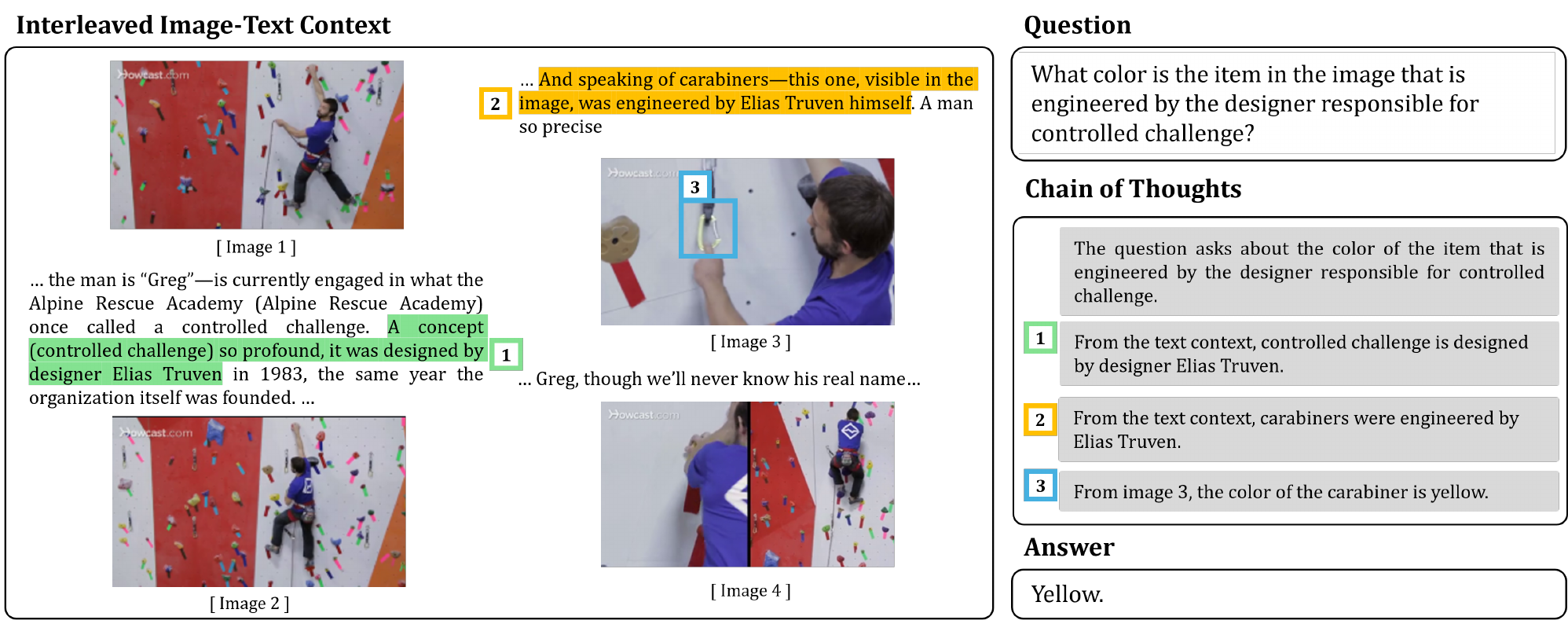}
\captionof{figure}{
\textbf{Qualitative Example of CRIT in Video Frame Domain.}
}
\label{fig:crux_example_1}
\end{figure*}

\begin{figure*}[t]
\centering
\includegraphics[width=1\textwidth]{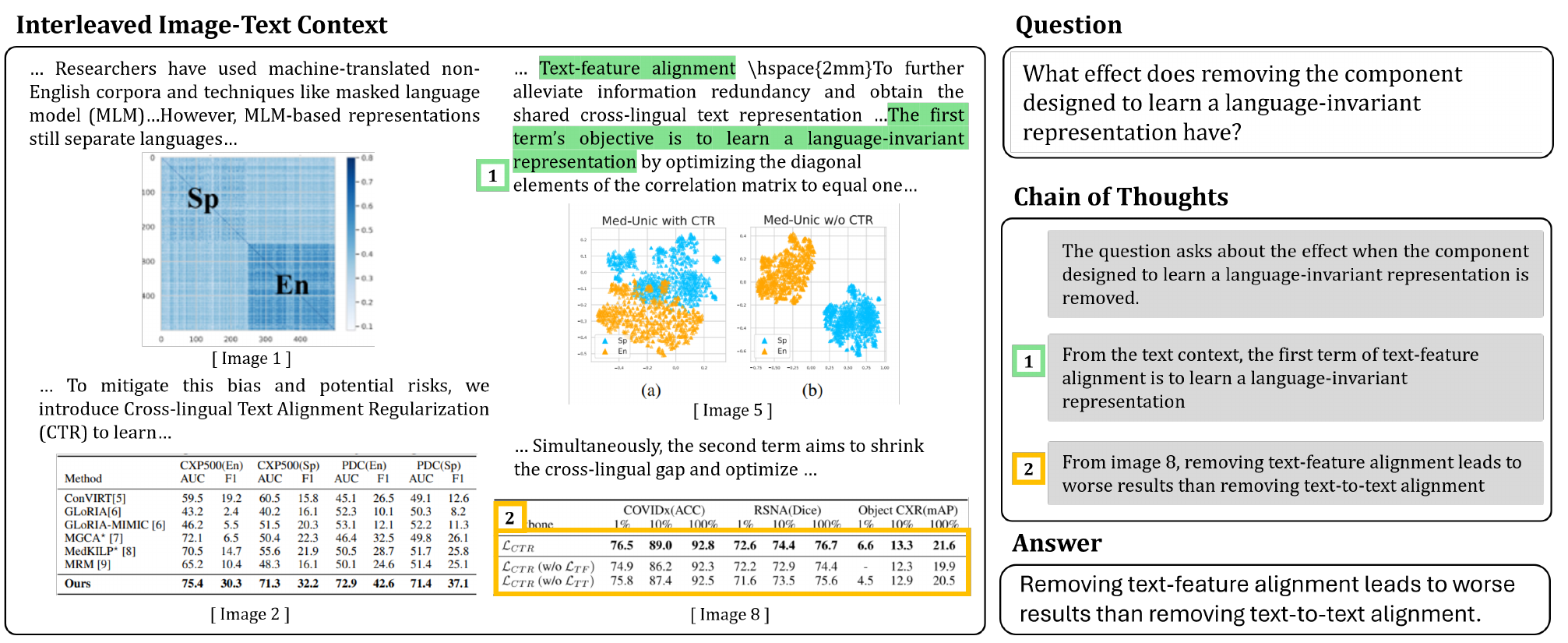}
\captionof{figure}{
\textbf{Qualitative Example of CRIT in Scientific Paper Domain.}
}
\label{fig:crux_example_2}
\end{figure*}

%% file: figures/prompt_w_wo_graph_compare.tex
\begin{figure*}[t]
\begin{tcolorbox}[
    colframe=gray!50!black, 
    colback=gray!10!white, 
    title=VLM-as-a-Judge Prompt for Pairwise QA Evaluation, 
    width=\textwidth, sharp corners=southwest, 
    left=1mm, right=1mm, top=0mm, bottom=0mm
]
\scriptsize
\begin{lstlisting}[
    aboveskip=0pt,
    belowskip=0pt,
    breakatwhitespace=false,
    basicstyle=\ttfamily\scriptsize,
    showstringspaces=false,
    frame=none
]
You are a VLM judge comparing TWO question-answer (QA) pairs that share the SAME images
and the SAME textual context, designed for evaluating cross-modal multi-hop reasoning.

You are given:
- One or more images (shared by both QA pairs).
- A single shared textual context.
- QA Pair 1: a question and its answer.
- QA Pair 2: a question and its answer.

Your task:
- Compare the TWO QA pairs directly.
- Decide which QA pair is structurally better for evaluating cross-modal multi-hop reasoning.

EVALUATION CRITERIA

1. Multi-hop and modality use
- Requires combining multiple pieces of evidence.
- Images contribute essential information (not decorative).
- One modality alone is insufficient.
- Should ask about entities, objects, or attributes that can only be discovered by inspecting the image.
- Does not leak visual discoveries in the text or question.
- Does not explicitly name visual attributes the solver should discover.
- Compound questions should be avoided that bundle multiple sub-questions with loosely related reasoning paths.

2. Question structure and clarity
- Precise and unambiguous.
- Avoids revealing intermediate reasoning steps or guarantees.
- Maintains a single coherent reasoning chain.

3. Evidence sufficiency and answer validity
- All required information is present in the image(s) and text.
- The answer is correct and supported.
- No outside knowledge is required.

DECISION RULES
- Output 1 if QA Pair 1 is clearly better.
- Output 2 if QA Pair 2 is clearly better.
- Output "tie" if both are roughly equivalent.

OUTPUT FORMAT

Return EXACTLY one JSON object and nothing else:

{
"decision": 1 | 2 | "tie",
"justification": "2-4 sentences explaning why this QA pair is better or why they are equivalent, focusing on 
structure, modality usage, and leakage."
}

Text Context:
{shared_context}

====================

QA PAIR 1
Question:
{q1}
Answer:
{a1}

--------------------

QA PAIR 2
Question:
{q2}
Answer:
{a2}

\end{lstlisting}
\end{tcolorbox}
\caption{VLM-as-a-Judge Prompt for Pairwise QA Evaluation.}
\label{fig:prompt_judge_compare}
\end{figure*}

%% file: figures/prompt_graph_construction.tex
\begin{figure*}[t]
% \begin{promptbox}{API call Generation Prompt - 1. First Call}
\begin{tcolorbox}[
    colframe=gray!50!black, 
    colback=gray!10!white, 
    title=Text Node Generation Prompt Type 1, 
    width=\textwidth, sharp corners=southwest, 
    left=1mm, right=1mm, top=0mm, bottom=0mm
]
\scriptsize
\begin{lstlisting}[
    aboveskip=0pt,
    belowskip=0pt,
    breakatwhitespace=false,
    basicstyle=\ttfamily\scriptsize,
    showstringspaces=false,
    frame=none
]
You are generating a fact about an object or entity that appears in an image.

Inputs you will receive:
- Object: the target entity.
- Image Caption: a one-sentence description of the whole image involving the object/entity.
- Object Caption: a short description focusing specifically on the object/entity.

Task:
Generate exactly one NEW fact about the object/entity in the category: Authorship / Creation
/ Discovery.

Rules:
- The fact must be NON-VISUAL (cannot be inferred from appearance or caption).
- The fact must be NON-COMMONSENSE (not universally true or obvious).
- Do not contradict either caption.
- Avoid mythical, fantasy, or obviously fictional names, rituals, or events.
- Names can be synthetic but should sound plausible.

Output format (JSON only):
{{
"subject": "<the given object/entity>",
"relation": "<the relation type>",
"object": "<the new entity, formatted as 'type (name)'>"
}}

Examples:
- {{"subject": "chair", "relation": "designed by", "object": "artisan (Liora Vex)"}}
- {{"subject": "compass", "relation": "invented by", "object": "engineer (Tavian Sorrell)"}}
- {{"subject": "man", "relation": "discovered", "object": "invention (Quantum Lens)"}}

Object: {object}
Image Caption: {image_caption}
Object Caption: {object_caption}
\end{lstlisting}
\end{tcolorbox}
\caption{Text Node Generation Prompt Type 1.}
\label{fig:tng_prompt_1}
\end{figure*}
\clearpage
\begin{figure*}[t]
% \begin{promptbox}{API call Generation Prompt - 1. First Call}
\begin{tcolorbox}[
    colframe=gray!50!black, 
    colback=gray!10!white, 
    title=Text Node Generation Prompt Type 2, 
    width=\textwidth, sharp corners=southwest, 
    left=1mm, right=1mm, top=0mm, bottom=0mm
]
\scriptsize
\begin{lstlisting}[
    aboveskip=0pt,
    belowskip=0pt,
    breakatwhitespace=false,
    basicstyle=\ttfamily\scriptsize,
    showstringspaces=false,
    frame=none
]
You are generating a fact about an object or entity that appears in an image.

Inputs you will receive:
- Object: the target entity.
- Image Caption: a one-sentence description of the whole image involving the object/entity.
- Object Caption: a short description focusing specifically on the object/entity.

Task:
Generate exactly one NEW fact about the object/entity in the category: 
Human Involvement / Institutional Association.

Rules:
- The fact must be NON-VISUAL (cannot be inferred from appearance or caption).
- The fact must be NON-COMMONSENSE (not universally true or obvious).
- Do not contradict either caption.
- Avoid mythical, fantasy, or obviously fictional names, rituals, or events.
- Names can be synthetic but should sound plausible.

Output format (JSON only):
{{
"subject": "<the given object>",
"relation": "<the relation type>",
"object": "<the new entity, formatted as 'type (name)'>"
}}

Examples:
- {{"subject: "man", "relation": "employed by", "object": "company (TechNova)"}}
- {{"subject: "boy", "relation": "friend of", "object": "person (Elias Thorn)"}}

Object: {object}
Image Caption: {image_caption}
Object Caption: {object_caption}
\end{lstlisting}
\end{tcolorbox}
\caption{Text Node Generation Prompt Type 2.}
\label{fig:tng_prompt_2}
\end{figure*}

\begin{figure*}[t]
% \begin{promptbox}{API call Generation Prompt - 1. First Call}
\begin{tcolorbox}[
    colframe=gray!50!black, 
    colback=gray!10!white, 
    title=Text Node Generation Prompt Type 3, 
    width=\textwidth, sharp corners=southwest, 
    left=1mm, right=1mm, top=0mm, bottom=0mm
]
\scriptsize
\begin{lstlisting}[
    aboveskip=0pt,
    belowskip=0pt,
    breakatwhitespace=false,
    basicstyle=\ttfamily\scriptsize,
    showstringspaces=false,
    frame=none
]
You are generating a fact about an object or entity that appears in an image.

Inputs you will receive:
- Object: the target entity.
- Image Caption: a one-sentence description of the whole image involving the object/entity.
- Object Caption: a short description focusing specifically on the object/entity.

Task:
Generate exactly one NEW fact about the object/entity in the category: 
Temporal / Historical Facts.

Rules:
- The fact must be NON-VISUAL (cannot be inferred from appearance or caption).
- The fact must be NON-COMMONSENSE (not universally true or obvious).
- Do not contradict either caption.
- Avoid mythical, fantasy, or obviously fictional names, rituals, or events.
- Names can be synthetic but should sound plausible.

For years/ages, make them synthetic but plausible (e.g., "year (2005)", "4 years").
Every object exists in the current year so the year or age should not be out of
a reasonable range.

Output format (JSON only):
{{
"subject": "<the given object>",
"relation": "<the relation type>",
"object": "<the new entity, formatted as 'type (name)'>"
}}

Examples:
- {{"subject": "car", "relation": "manufactured in", "object": "year (2005)"}}
- {{"subject": "dog", "relation": "has age", "object": "4 years"}}
- {{"subject": "spoon", "relation": "made in", "object": "year (2010)"}}

Object: {object}
Image Caption: {image_caption}
Object Caption: {object_caption}
\end{lstlisting}
\end{tcolorbox}
\caption{Text Node Generation Prompt Type 3.}
\label{fig:tng_prompt_3}
\end{figure*}

\begin{figure*}[t]
% \begin{promptbox}{API call Generation Prompt - 1. First Call}
\begin{tcolorbox}[
    colframe=gray!50!black, 
    colback=gray!10!white, 
    title=Edge Generation Prompt, 
    width=\textwidth, sharp corners=southwest, 
    left=1mm, right=1mm, top=0mm, bottom=0mm
]
\scriptsize
\begin{lstlisting}[
    aboveskip=0pt,
    belowskip=0pt,
    breakatwhitespace=false,
    basicstyle=\ttfamily\scriptsize,
    showstringspaces=false,
    frame=none
]
You are given a list of entities in the format "type (name)".  
Your task is to generate plausible synthetic relations between these entities.

### Rules:
- Each output must be a JSON object with the format:
{{
   "subject": "<entity from the list>",
   "relation": "<synthetic relation connecting it to another entity>",
   "object": "<entity from the list>"
}}

- The 'subject' and 'object`' must always come from the given list.  
- Relations must be plausible. If there is no reasonable relation, the output should be
an empty list.
- Do not invent new entities outside of the given list.  
- Output only a list of JSON objects (no extra text).

### Example:
Input:
["institution (Museum of Oracles)", "event (Expo 2020)", "artifact (Singing Blade)",
"concept (Eternal Silence)"]

Output:
[
   {{"subject": "institution (Museum of Oracles)", "relation": "preserves",
     "object": "artifact (Singing Blade)"}},
   {{"subject": "concept (Eternal Silence)", "relation": "inspires", 
     "object": "artifact (Singing Blade)"}},
   {{"subject": "event (Expo 2020)", "relation": "hosts", 
     "object": "institution (Museum of Oracles)"}},
   {{"subject": "artifact (Singing Blade)", "relation": "represents", 
     "object": "concept (Eternal Silence)"}},
   {{"subject": "institution (Museum of Oracles)", "relation": "exhibits", 
     "object": "event (Expo 2020)"}},
]

### Input:
{list_of_entities}

### Output:

\end{lstlisting}
\end{tcolorbox}
\caption{Edge Generation Prompt.}
\label{fig:prompt_graph_construction}
\end{figure*}
\clearpage

%% file: figures/prompt_textual_context_generation.tex
\begin{figure*}[t]
% \begin{promptbox}{API call Generation Prompt - 1. First Call}
\begin{tcolorbox}[
    colframe=gray!50!black, 
    colback=gray!10!white, 
    title=Textual Context Generation Prompt, 
    width=\textwidth, sharp corners=southwest, 
    left=1mm, right=1mm, top=0mm, bottom=0mm
]
\scriptsize
\begin{lstlisting}[
    aboveskip=0pt,
    belowskip=0pt,
    breakatwhitespace=false,
    basicstyle=\ttfamily\scriptsize,
    showstringspaces=false,
    frame=none
]
You are writing a {context_type}.

The following entities and relations must be included:

Entities:
[Entities list]

Relations:
[Relations list]

Detailed Guidelines:
1. Explicit Image References
- Every entity that contains "(Image N)" MUST be explicitly tied to its image number in the
text. Do this by phrases like "as seen in image N", "shown in image N", or "visible in
image N".  
- Every entity that contains "(Image)" MUST be described as appearing in that image. Do this
by phrases like "as seen in the image", "shown in the image", or "visible in the image".
- Example: Instead of writing "The telephone pole is maintained by Veridian Grid Solutions",
write "The telephone pole shown in image 1 is maintained by Veridian Grid Solutions."

2. Inclusion of All Entities & Relations 
- Every entity listed above MUST appear in the generated text.  
- Every relation MUST be expressed clearly, connecting the subject and object naturally.  
- You may rephrase the relation semantically, but the meaning must remain intact.

3. Integration into Natural Writing 
- Blend the entities and relations into a flowing narrative appropriate for the chosen
context type ({context_type}).  
- Avoid bullet-point style in the output; it must read like a coherent piece of writing.  
- The writing should be creative but faithful to the factual structure provided.

4. No Contradictions or New Visual Details 
- Do NOT invent or assign new visual attributes to entities (e.g., do not say "the pole is
red" if not given).  
- You may add context, background, or imaginative framing, but it must not contradict the
given information.

5. Optional Creative Expansion  
- You may enrich the text with atmosphere, tone, or style fitting the chosen context type.
- Added information must support, not override, the provided facts

\end{lstlisting}
\end{tcolorbox}
\caption{Textual Context Generation Prompt.}
\label{fig:prompt_textual_context_generation}
\end{figure*}

%% file: figures/prompt_question_answer_pair_generation.tex
\begin{figure*}[t]
% \begin{promptbox}{API call Generation Prompt - 1. First Call}
\begin{tcolorbox}[
    colframe=gray!50!black, 
    colback=gray!10!white, 
    title=Question-Answer Pair Generation Prompt, 
    width=\textwidth, sharp corners=southwest, 
    left=1mm, right=1mm, top=0mm, bottom=0mm
]
\scriptsize
\begin{lstlisting}[
    aboveskip=0pt,
    belowskip=0pt,
    breakatwhitespace=false,
    basicstyle=\ttfamily\scriptsize,
    showstringspaces=false,
    frame=none
]
You are given a list of structured graph triples sampled from a graph. Each triple is a
JSON object with the keys "subject", "relation", and "object". Your task is to generate a
multi-step question that requires reasoning across ALL the provided triples step-by-step.
The question must not be answerable using only a subset of the triples.

Guidelines:
- The question must not mention any entities that should be inferred. All the intermediate
entities should be inferred step-by-step.
- The final answer to the question must be the last "object" entity in the last triple
- Always mention the image reference in the question if it exists 
(e.g., "object in the image").
- Break the question into two sentences if it is too long or complex to keep it clear and
understandable in one sentence. The second sentence should add new context, not repeat
the same information from the first.

Generate a multi-step question and answer, and respond with ONLY a valid JSON object in the
following format:
{{
  "question": "...",
  "answer": "..."
}}
  
Triples:
[
  {{"subject": "blue", "relation": "is the color of", "object": "cord (Image)"}},
  {{"subject": "cord", "relation": "used during", "object": "event (Product Launch Demo)"}},
  {{"subject": "event (Product Launch Demo)", "relation": "event (Product Launch Demo)
     hosts utility company (Veridian Grid Solutions)", "object": "utility company (Veridian
     Grid Solutions)"}},
  {{"subject": "utility company (Veridian Grid Solutions)", "relation": "utility company
     (Veridian Grid Solutions) maintained by telephone pole", "object": "telephone pole"}},
  {{"subject": "telephone pole (Image)", "relation": "is", "object": "black"}}
]
Notes:
- The answer must be "black" because it is the last object in the last triple.
- The following entities should not be mentioned directly in the question as they
are inferred step-by-step: cord, event (Product Launch Demo), utility company (Veridian
Grid Solutions), telephone pole.
Output:
{{
  "question": "What is the color of the object in the image that maintains the company that
  hosts the event, where the event uses a blue object that is to the left of camera?",
  "answer": "black"
}}

Triples:
[
  {{"subject": "research team (Savanna Ecology Project)", "relation": "research team
     (Savanna Ecology Project) studied giraffe", "object": "giraffe"}},
  {{"subject": "giraffe (Image)", "relation": "is", "object": "walking"}}
]
Notes:
- The answer must be "walking" because it is the last object in the last triple.
- The following entity should not be mentioned directly in the question as it is inferred
step-by-step: giraffe.
Output:
{{
  "question": "What is the entity in the image doing that is studied by the research team
  known as the Savanna Ecology Project?",
  "answer": "walking"
}}

Triples:
[
  {triples}
]
The answer must be "{last_object}" because it is the last object in the last triple.
The following entities should not be mentioned directly in the question as they are inferred
step-by-step: {', '.join(intermediate_objects)}.
Output:

\end{lstlisting}
\end{tcolorbox}
\caption{Question-Answer Pair Generation Prompt.}
\label{fig:prompt_question_answer_pair_generation}
\end{figure*}

%% file: figures/prompt_cot_response_generation.tex
\begin{figure*}[t]
% \begin{promptbox}{API call Generation Prompt - 1. First Call}
\begin{tcolorbox}[
    colframe=gray!50!black, 
    colback=gray!10!white, 
    title=CoT Response Generation Prompt, 
    width=\textwidth, sharp corners=southwest, 
    left=1mm, right=1mm, top=0mm, bottom=0mm
]
\scriptsize
\begin{lstlisting}[
    aboveskip=0pt,
    belowskip=0pt,
    breakatwhitespace=false,
    basicstyle=\ttfamily\scriptsize,
    showstringspaces=false,
    frame=none
]
You are given a question, its correct answer, and a subgraph that contains the entities and
relations supporting the QA pair. 
Your task is to generate a detailed chain-of-thought reasoning output that explains step by
step how the answer follows from the question. 

Requirements for the reasoning:
1. Explicitly mention the source of each piece of information:
  - If the evidence comes from an image, say "from image X".  
  - If the evidence comes from a figure/table, say "from figure/table Y".  
  - If no image or figure/table is involved, assume the information is from the text context
    and say "from the text context".  
2. Trace through the relevant entities and relations in the subgraph in logical order.  
3. End with the conclusion that matches the provided answer.  
4. The reasoning should read naturally, as if another model is thinking through the problem
   step by step.
5. Assume the reader is looking at the images/figures/tables and the text context to answer
   the question.
6. The subgraph is only for reference. The actual reader will not see the subgraph so don't
   generate as if the reader is seeing the subgraph. Don't say anything like "from the
   subgraph", "the relation shows", or "the entity indicates".
7. Do not generate any unnecessary reasoning steps that repeat the same information which is
   already mentioned in previous steps.

Question: What action is performed by individual trained at the institution in the image?
Answer: continues dancing around room
Subgraph: [
{{'subject': 'institution (Central Academy of Contemporary Movement)',
  'object': 'young woman',
  'relation': 'young woman trained under institution
              (Central Academy of Contemporary Movement)'}},
{{'subject': 'young woman',
  'object': None,
  'relation': 'dancing around room',
  'image': 'image 3'}}]
  
Chain-of-thought reasoning:
The question asks what action is performed by the person trained at the institution. 
From the text context, the institution is the Central Academy of Contemporary Movement,
and a young woman trained there. 
From image 3, I can see the woman is dancing around the room.
Therefore, the action performed is dancing around room.

Question: What method has lower time cost compared to the another method that is based on an
          algorithm used to obtain the traditional CVT through iterative updates until
          convergence?
Answer: time cost
Subgraph: [
{{'source_entity': "Lloyd's algorithm",
  'target_entity': 'CVT',
  'relationship_description': "The traditional CVT is usually obtained by Lloyd's algorithm,
  iteratively performing updates after each assignment step until convergence is reached."}},
{{'source_entity': 'SLIC',
  'target_entity': "Lloyd's algorithm",
  'relationship_description': "SLIC generates superpixels based on Lloyd's algorithm"}},
{{'source_entity': 'SLIC',
  'target_entity': 'FLIC',
  'relationship_description': 'FLIC's time cost is lower than SLIC's time cost',
  'figure': 'Figure 4'}}]
  
Chain-of-thought reasoning:
The question asks about what method has lower time cost compared to another method based on
an algorithm for computing the traditional CVT.
From the text context, the traditional CVT is obtained by Lloyd's algorithm, which
iteratively updates until convergence. 
From the text context, the method SLIC is based on Lloyd's algorithm. 
From Figure 4, it is shown that FLIC's time cost is lower than SLIC's time cost.
Therefore, the method with lower time cost is FLIC.

Question: {question}
Answer: {answer}
Subgraph: {subgraph}
Chain-of-thought reasoning:

\end{lstlisting}
\end{tcolorbox}
\caption{CoT Response Generation Prompt.}
\label{fig:prompt_cot_response_generation}
\end{figure*}

%% file: figures/prompt_caption_to_graph.tex
\begin{figure*}[t]
% \begin{promptbox}{API call Generation Prompt - 1. First Call}
\begin{tcolorbox}[
    colframe=gray!50!black, 
    colback=gray!10!white, 
    title=Video Caption to Graph Conversion Prompt, 
    width=\textwidth, sharp corners=southwest, 
    left=1mm, right=1mm, top=0mm, bottom=0mm
]
\scriptsize
\begin{lstlisting}[
    aboveskip=0pt,
    belowskip=0pt,
    breakatwhitespace=false,
    basicstyle=\ttfamily\scriptsize,
    showstringspaces=false,
    frame=none
]
You will be given a list of captions (one per scene, in order).
Each caption is paired with a time range.

Your job is to produce BOTH a global entity inventory and per-scene graphs, in two separate
sections.

REQUIREMENTS:

1. Entities section:
- Deduplicate entities across all captions into a single global inventory.
- Assign IDs as stringified integers ("1","2","3").
- Always include "attributes": [].
- Attributes = static/descriptive properties (e.g., "red", "wooden", "wearing hat").
- Do not include transient states (e.g., "sitting", "throwing") here.
- If multiple similar entities are indistinguishable -> group them (e.g., "two dogs").
- If entities are clearly distinct -> create differentiated forms (e.g., "bag_1", "bag_2").
- Do not include interactions with other entities here.
- IMPORTANT: If two entity mentions occur in overlapping or adjacent time ranges, they are
considered coreference *candidates*. 
    - Merge them into the same global entity only if the semantics clearly indicate they are
    the same entity 
    - (e.g., "former president" at 3s-10s and "man gives a speech" at 5s-12s).

2. Scenes section:
- Each caption corresponds to a scene labeled "Scene N" with its time range (use the given
order, even if time ranges overlap).
- Each scene has a key "relations".
- "relations" is a list of relation triples for that scene.
- Each relation triple must have:
    - "source": source entity ID
    - "target": target entity ID OR null (if no second entity is involved)
    - "relation": a **brief phrase** (not a full sentence) that concisely describes the
    action or interaction 
    (e.g., "dog chases cat", "man hands bag to woman", "gives a speech").
- A relation exists if:
    - At least two distinct entities interact, OR
    - A single entity performs an action (then target = null).
- If no valid relations exist for a scene, output "relations": [].

3. General:
- Separate the "entities" section from the "scenes" section.
- Keep relations directional and minimal; avoid redundant inverses.
- Do not invent entities not grounded in captions.

OUTPUT FORMAT:

{{
"entities": [
    {{"id":"1","entity":"ENTITY NAME","attributes":[]}}
],
"scenes": [
    {{
    "scene":"scene_1",
    "relations":[
{{"source":"1","target":"2","relation":"brief phrase describing interaction between entity 1
   and entity 2"}},
{{"source":"1","target":null,"relation":"solo action"}}
    ]
    }}
]
}}

INPUT CAPTIONS:
{annotated}

\end{lstlisting}
\end{tcolorbox}
\caption{Video Caption to Graph Conversion Prompt.}
\label{fig:prompt_caption_to_graph}
\end{figure*}

%% file: figures/prompt_paper_nonfig_paragraph_to_graph.tex
\begin{figure*}[t]
% \begin{promptbox}{API call Generation Prompt - 1. First Call}
\begin{tcolorbox}[
    colframe=gray!50!black, 
    colback=gray!10!white, 
    title=Paragraph(without figure reference) to Graph Transformation Prompt, 
    width=\textwidth, sharp corners=southwest, 
    left=1mm, right=1mm, top=0mm, bottom=0mm
]
\scriptsize
\begin{lstlisting}[
    aboveskip=0pt,
    belowskip=0pt,
    breakatwhitespace=false,
    basicstyle=\ttfamily\scriptsize,
    showstringspaces=false,
    frame=none
]
 -Goal-
Given a scientific text and a list of scientific entities, identify ALL relations 
expressed in the paragraph.

-Steps-
1. Use the paragraph text only (no figures/tables).
2. Extract any relation between entities that is explicitly stated in the text.
3. For each valid relation, output a JSON object with:
   - source_entity (must come from the provided entities list)
   - target_entity (must come from the provided entities list, or null if not applicable)
   - relationship_description (short, factual, and grounded in the text)
4. Do NOT invent entities that are not in the input entities list. 
   Do NOT invent relations and do NOT use prior knowledge.
5. Output must be valid JSON as a list of relation objects (not wrapped in another key).
6. Each relation dictionary must be one line (compact JSON style).
7. Do not include any text outside the JSON.
8. If there are no relations, return an empty list: []

Output Format Example:
[
  {{"source_entity": "Entity1", "target_entity": "Entity2", "relationship_description":
    "Description of relation grounded in text"}},
  {{"source_entity": "Entity3", "target_entity": null, "relationship_description":
    "Another relation grounded in text"}}
]

######################
Entities:
{entities_json}

Referenced Paragraph:
{paragraph}

######################
Output (JSON only):

\end{lstlisting}
\end{tcolorbox}
\caption{Paragraph(without figure reference) to Graph Transformation Prompt.}
\label{fig:prompt_paper_nonfig_paragraph_to_graph}
\end{figure*}

%% file: figures/prompt_paper_fig_paragraph_to_graph.tex
\begin{figure*}[t]
% \begin{promptbox}{API call Generation Prompt - 1. First Call}
\begin{tcolorbox}[
    colframe=gray!50!black, 
    colback=gray!10!white, 
    title=Paragraph(with figure reference) to Graph Transformation Prompt, 
    width=\textwidth, sharp corners=southwest, 
    left=1mm, right=1mm, top=0mm, bottom=0mm
]
\scriptsize
\begin{lstlisting}[
    aboveskip=0pt,
    belowskip=0pt,
    breakatwhitespace=false,
    basicstyle=\ttfamily\scriptsize,
    showstringspaces=false,
    frame=none
]
 -Goal-
Given a scientific text (with one or more figure/table captions and references) 
and a list of scientific entities, identify ONLY the relations that are 
explicitly supported by the figures/tables.

-Steps-
1. Use both the paragraph text and the provided figures/tables.
2. Extract a relation ONLY IF it is directly grounded in the figure or table:
  - Numerical results or metrics reported in the figure/table. Include numerical values if
  available.
  - Explicit comparisons shown in the figure/table (e.g., "X outperformed Y").
  - Visual descriptors of figure elements (e.g., "yellow line corresponds to Model A").
3. Ignore the following completely:
  - Interpretations, hypotheses, or explanations
  (e.g., "improvements are due to larger size").
  - Background details (methods, datasets, phases, tasks, architectures).
4. For each valid relation, output a JSON object with:
   - source_entity (must come from the provided entities list)
   - target_entity (must come from the provided entities list, or null if not applicable)
   - relationship_description (short, factual, and grounded in the figure/table)
   - figure (the figure_label where the relation is supported, never null)
   - idx (list of one or more sentence indices where relation appears, e.g. [0] or [1,2])
5. Do NOT invent entities that are not in the input entities list. Do NOT invent relations
   and do NOT use prior knowledge.
6. Output must be valid JSON as a list of relation objects (not wrapped in another key).
7. Each relation dictionary must be one line (compact JSON style).
8. Do not include any text outside the JSON.
9. If there are no figure/table-grounded relations, return an empty list: []

Output Format Example:
[
  {{"source_entity": "Entity1", "target_entity": "Entity2", "relationship_description":
    "Description of relation grounded in figure/table", "figure": "FigureX", "idx": [0]}},
  {{"source_entity": "Entity3", "target_entity": null, "relationship_description":
    "Another relation grounded in figure/table", "figure": "TableY", "idx": [1,2]}}
]

######################
Figures:
{figures_json}

Entities:
{entities_json}

Indexed Paragraph Sentences:
{sentences_text}

######################
Output (JSON only):

\end{lstlisting}
\end{tcolorbox}
\caption{Paragraph(with figure reference) to Graph Transformation Prompt.}
\label{fig:prompt_paper_fig_paragraph_to_graph}
\end{figure*}